\documentclass[journal,twoside,web]{ieeecolor}
\usepackage{tmi}
\usepackage{cite}
\usepackage{amsmath,amssymb,amsfonts}
\usepackage{algorithmic}
\usepackage{graphicx}
\usepackage{textcomp}
\usepackage{amssymb, stmaryrd}   
\usepackage{wrapfig}
\usepackage{bm}
\usepackage{newclude}
\usepackage{bm}
\usepackage{booktabs}
\usepackage{array} 
\usepackage{xcolor}
\usepackage{multirow}
\usepackage{subcaption}
\usepackage{caption}
\usepackage[table]{xcolor}
\usepackage{soul} 
\definecolor{citeblue}{RGB}{48,111,186}

\usepackage[pagebackref=false,breaklinks=true,colorlinks=true,citecolor=citeblue,bookmarks=false]{hyperref}
\usepackage{tcolorbox}

\definecolor{bestcolor}{rgb}{0.95, 0.0, 0.0}
\definecolor{secondcolor}{rgb}{0.0, 0.0, 0.8}
\definecolor{mydarkgreen}{RGB}{0, 100, 0}

\newcommand{\best}[1]{{\color{bestcolor}{{#1}}}}
\newcommand{\second}[1]{\color{secondcolor}{#1}}

\def\BibTeX{{\rm B\kern-.05em{\sc i\kern-.025em b}\kern-.08em
    T\kern-.1667em\lower.7ex\hbox{E}\kern-.125emX}}
\begin{document}
\title{Medical Knowledge Intervention Prompt Tuning for Medical Image Classification}
\author{Ye Du, Nanxi Yu, and Shujun Wang, \IEEEmembership{Member, IEEE}
\thanks{This work was partially supported by RGC Collaborative Research Fund (No. C5055-24G), the Start-up Fund of The Hong Kong Polytechnic University (No. P0045999), the Seed Fund of the Research Institute for Smart Ageing (No. P0050946), and Tsinghua-PolyU Joint Research Initiative Fund (No. P0056509), and PolyU UGC funding (No. P0053716).}
\thanks{Ye Du and Nanxi Yu are with the Department of Biomedical Engineering, The Hong Kong Polytechnic University, Hong Kong SAR, China. (E-mail: \{duyee.du, nx-nancy.yu\}@connect.polyu.hk)}
\thanks{Shujun Wang is with the Department of Biomedical Engineering, The Hong Kong Polytechnic University, Hong Kong SAR, China. Shujun Wang is also affiliated with Research Institute for Smart Ageing, The Hong Kong Polytechnic University, Hong Kong SAR, China. Shujun Wang is also affiliated with Research Institute for Artificial Intelligence of Things, The Hong Kong Polytechnic University, Hong Kong SAR, China. (E-mail: shu-jun.wang@polyu.edu.hk)} 
\thanks{Shujun Wang is the corresponding author.}
}

\maketitle

\begin{abstract}
Vision-language foundation models (VLMs) have shown great potential in feature transfer and generalization across a wide spectrum of medical-related downstream tasks. 
However, fine-tuning these models is resource-intensive due to their large number of parameters.
Prompt tuning has emerged as a viable solution to mitigate memory usage and reduce training time while maintaining competitive performance. 
%
Nevertheless, the challenge is that existing prompt tuning methods cannot precisely distinguish different kinds of medical concepts, which miss essentially specific disease-related features {across various medical imaging modalities in medical image classification tasks}.
We {find} that Large Language Models (LLMs), trained on extensive text corpora, are particularly adept at providing this specialized medical knowledge. 
Motivated by this, we propose incorporating LLMs into the prompt tuning process. Specifically, we introduce the CILMP, Conditional Intervention of Large Language Models for Prompt Tuning, a method that bridges LLMs and VLMs to facilitate the transfer of medical knowledge into VLM prompts. 
CILMP extracts disease-specific representations from LLMs, intervenes within a low-rank linear subspace and utilizes them to create disease-specific prompts. 
Additionally, a conditional mechanism is incorporated to condition the intervention process on each individual medical image, generating instance-adaptive prompts and thus enhancing adaptability.
Extensive experiments across diverse medical image datasets demonstrate that CILMP consistently outperforms state-of-the-art prompt tuning methods, demonstrating its effectiveness.
{Code is available at} \href{https://github.com/usr922/cilmp}{https://usr922.github.io/cilmp}.

\end{abstract}
\vspace{0.5em}


\section{Introduction}\label{sec:intro}
\IEEEPARstart{W}{ith}
the rapid advancement of deep learning technologies, numerous studies have been proposed to enhance computer-aided diagnosis within medical data analysis. 
Recently, vision-language foundation models (VLMs) \cite{radford2021learning,jia2021scaling,alayrac2022flamingo,li2022blip,yucoca,yang2023attentive,huang2023visual,zhang2023knowledge,kim2024transparent,christensen2024vision}, which are trained on extremely large-scale data encompassing multiple domains and data distributions, have demonstrated significant potential in feature transfer and generalization across a wide array of downstream tasks. 
However, the fine-tuning of these foundation models is often prohibitively expensive due to the substantial number of model parameters involved.

To address the above challenge, parameter efficient fine-tuning (PEFT) methods \cite{zhou2022learning, hulora, zhou2022conditional, gao2024clip, ding2023parameter, wu2024reft, han2024parameter} have emerged as a promising alternative by updating only a small subset of the model's weights while achieving performance levels comparable to full fine-tuning in many scenarios \cite{khattak2023self, duttparameter, zheng2024large, roy2024consistency, zhou2023zegclip, li2024cascade, lian2024less}.
%
One of the most effective PEFT methods for adapting VLMs to downstream tasks is prompt tuning \cite{zhou2022learning, zhou2022conditional}. 
This technique, which has its origins in the field of natural language processing \cite{li2021prefix, zhong2021factual, lester2021power, he2022hyperprompt, liu2023pre}, involves maintaining the fixed parameters of the large model while training a set of learnable tokens, referred to as prompt contexts \cite{zhou2022learning}. 
These prompt contexts are concatenated with class names to serve as the input to the text encoder within VLMs. 
Following optimization, these prompts are employed to generate cosine classifiers~\cite{gidaris2018dynamic,zhou2022learning} for each category, thereby facilitating a range of tasks such as natural image classification \cite{yao2023visual, lee2023read, yao2024tcp, kim2024aapl, zhang2024dept, bulat2023lasp, khattak2023self, zheng2024large, roy2024consistency} and semantic segmentation \cite{zhou2023zegclip, zhu2023segprompt, li2024cascade}. 
The straightforward nature of this paradigm highlights its potential to exploit the capabilities of pre-trained VLMs while preserving flexibility and achieving high performance.

Despite its success in natural image domains, the application of prompt tuning in medical image analysis still faces challenges.
One primary challenge in medical imaging tasks is the requirement for precise differentiation of various medical concepts to accurately understand diseases, which underscores the necessity of integrating knowledge of the specialized medical domain \cite{xie2021survey, yang2019dscgans, zhang2023knowledge, bie2024xcoop}.
However, when adapting a pre-trained VLM to recognize ``basal cell carcinoma" from dermatoscopy images, conventional prompt tuning methods employ disease-agnostic learnable prompt context tokens.
These tokens lack the specific information necessary for the accurate identification of basal cell carcinoma, thereby constraining the full potential of the pre-trained VLM.
{Furthermore, existing prompt tuning methods typically construct shared prompt contexts across different categories.} \cite{zhou2022learning, zhou2022conditional, yao2023visual, lee2023read, bulat2023lasp, khattak2023self, cao2024domain, zhang2024dept}
%
{Such a design struggles with the fine-grained nature of medical image classification, as category names alone often do not provide sufficient information for recognition. }
For example, when distinguishing between ``viral pneumonia" and ``bacterial pneumonia" in X-ray images, the class names can barely provide adequate discriminative information to differentiate these two diseases due to their textual similarity.

Based on the preceding analysis, it is evident that incorporating medical domain knowledge into the prompt tuning process is essential.
Large language models (LLMs) \cite{touvron2023llama, touvron2023llama2, llama3modelcard, Chatgpt, anil2023palm, Gpt4, chiang2023vicuna}, which are trained on extensive text corpora, are particularly effective in providing this specialized knowledge \cite{zheng2024large, roy2024consistency, yang2023language}. 
These models possess a comprehensive understanding of various diseases, enabling them to generate context-specific information, discern subtle distinctions between similar medical conditions, and provide diverse medical knowledge. 
For instance, when queried about ``basal cell carcinoma", an LLM can offer detailed descriptions of its distinctive visual features like ``telangiectasia and rolled borders".
This professional and nuanced information can significantly enhance the richness and accuracy of the prompts, as it provides discriminative representations for each disease.

\begin{figure}[t]
\centering
\includegraphics[width=0.5\textwidth]{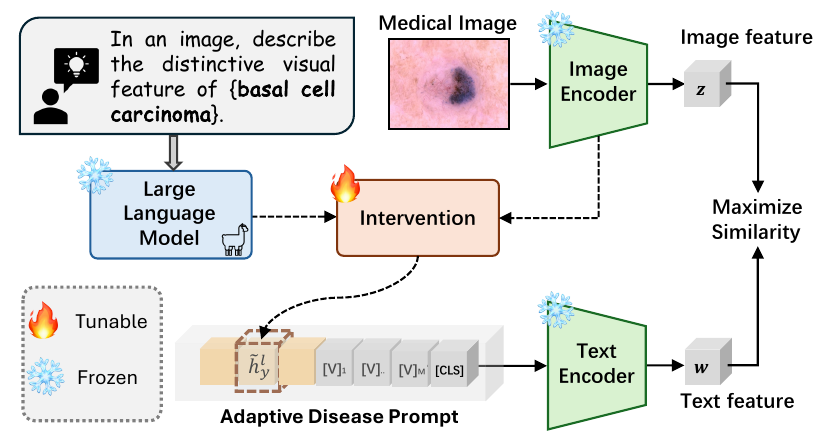}
\caption{{Concept illustration of our CILMP method. CILMP first extracts concept-aware representations from a frozen large language model. It then intervenes in these representations with the guidance of image features to generate the adaptive disease prompts for the VLM text encoder.}}
\label{fig:1}
\vspace{-1.5em}
\end{figure}

Motivated by these capabilities, this paper aims to integrate large language models into the prompt tuning process to enhance the adaptation of vision-language foundation models for medical image analysis. 
Specifically, we propose Conditional Intervention of Large Language Models for Prompt Tuning (CILMP), a framework designed to bridge the gap between LLMs and VLMs by facilitating the transfer of medical domain knowledge from LLMs into VLM prompts.
As shown in Fig. \ref{fig:1}, for a certain disease, our CILMP framework first extracts concept-aware representations from the LLM. 
Given that LLMs and VLMs are usually pre-trained on different data modalities and distributions, the CILMP framework mitigates this gap by intervening in the LLM representations within a low-rank linear subspace.
The corresponding intervention functions are trained to steer the framework's behavior towards accurate disease diagnosis, which is inspired by representation fine-tuning (ReFT) \cite{wu2024reft}.
%
%
Moreover, we propose learning the intervention functions guided by input medical images to generate instance-adaptive prompts.
This is achieved by conditioning the subspace intervention process on a relationship descriptor \cite{zhou2023zegclip, li2024cascade} between the image feature and the {non-intervened} LLM representations. 
Such a conditional mechanism allows for the incorporation of image-text matching prior into the prompts before making the final decision.
%
%
Notably, the CILMP framework requires only the learning of intervention functions applied to the LLM representations, without necessitating gradient flow through the LLM itself, thus preserving the efficiency of the prompt tuning process.
%

To evaluate the effectiveness of the proposed CILMP framework, we conduct extensive experiments across 11 diverse medical image datasets. 
These datasets encompass a wide range of imaging modalities, including dermatoscope, fundus, ultrasound, histopathology, and X-ray images, etc.
In our evaluation, 
{we compare CILMP against 15 recently proposed state-of-the-art prompt tuning methods} using four commonly used metrics: Accuracy, F1-score, Area Under Curve (AUC), and Kappa score. 
The experimental results demonstrate that CILMP consistently surpasses existing prompt tuning methods by considerable margins on 10 out of 11 datasets, thereby validating the effectiveness of our approach.

Our contributions are summarized as follows.
\begin{itemize}
\item We introduce the application of large language models to address the deficiencies in medical domain knowledge inherent in traditional prompt tuning methods for computer-aided diagnosis.
\item We propose CILMP as an intermediary framework that bridges LLMs and vision-language foundation models, facilitating the generation of class-specific and instance-adaptive prompts.
\item We perform extensive experiments to evaluate the performance of CILMP across a diverse set of datasets. Comparisons with recently proposed state-of-the-art prompt tuning methods demonstrate the effectiveness of our approach.
\end{itemize}

The rest of this paper is arranged as follows.
We delve into related works in Section~\ref{relatedwork}. We then elaborate on the technical details of our method in Section~\ref{sec:method}. Experimental results and discussions are presented in Section~\ref{sec:exp} and Section~\ref{sec:discuss}. Finally, we summarize the conclusions of this paper in Section~\ref{sec:conclusion}.

\section{Related Work}
\label{relatedwork}

\subsection{Large Language Models}

Recent advancements in pre-training LLMs \cite{touvron2023llama, touvron2023llama2, llama3modelcard, Chatgpt, anil2023palm, Gpt4, chiang2023vicuna} have significantly enhanced their ability to understand and generate human language, as evidenced by models such as ChatGPT \cite{Chatgpt}, GPT-4 \cite{Gpt4}, and LLaMA \cite{touvron2023llama}. 
These models, pre-trained on extensive corpora and comprising billions of parameters, excel in capturing linguistic patterns and contextual relationships. 
Notably, LLaMA~\cite{touvron2023llama} and its iterations, LLaMA-2 \cite{touvron2023llama2} and LLaMA-3 \cite{llama3modelcard}, have demonstrated improvements in scale, efficiency, and reasoning capabilities. 
Various studies have leveraged LLMs for downstream tasks, such as generating predefined prompts for image classification \cite{roy2024consistency}, proposing LLM-guided concept bottlenecks for interpretable classification \cite{yang2023language}, and jointly training LLMs and VLMs for classification tasks \cite{zheng2024large}. 
Despite their high performance, the training workload for these large models remains substantial.

\subsection{Vision Language Foundation Models}

Recent advancements have significantly bridged visual signals with linguistic semantics, particularly through the pre-training of VLMs~\cite{radford2021learning, jia2021scaling, alayrac2022flamingo, li2022blip, yucoca, yang2023attentive, huang2023visual}.
%
%
Notably, the CLIP model \cite{radford2021learning} has demonstrated effective zero-shot prediction capabilities using a contrastive learning objective by learning the correlation between image and text embeddings~\cite{oord2018representation}.
Following this paradigm, Jia \textit{et al.} \cite{jia2021scaling} achieved robust zero-shot inference with exascale noisy image alt-text data, while
%
Alayrac~\textit{et al.}~\cite{alayrac2022flamingo} bridge powerful pre-trained vision-only and language-only models to enable powerful in-context few-shot learning capabilities.
Li~\textit{et al.}~\cite{li2022blip} unified image-text understanding and generation, enhancing functionalities such as image captioning and visual question answering. 
%
Yang~\textit{et al.}~\cite{yang2023attentive} introduced an attentive masking mechanism to improve VLM pre-training efficiency.
%

In the medical imaging domain, notable efforts \cite{zhang2023knowledge, kim2024transparent, christensen2024vision} have focused on pre-training VLMs with large-scale medical data. 
For instance, Zhang~\textit{et al.}~\cite{zhang2023knowledge}~incorporated medical domain knowledge into pre-training for chest X-ray images using a knowledge graph.
%
Christensen \textit{et al.} \cite{christensen2024vision} developed a VLM for echocardiography, learning relationships between ultrasound images and expert interpretations.
%
Additionally, Kim \textit{et al.} \cite{kim2024transparent} highlighted VLMs' suitability for trustworthy and transparent medical AI systems.
%
%
For further details on VLMs, refer to \cite{zhang2024vision, shrestha2023medical}.

\vspace{-0.5em}
\subsection{Prompt Tuning}
%
%
%
%
Prompt tuning \cite{li2021prefix, zhong2021factual, lester2021power, he2022hyperprompt, liu2023pre, bie2024xcoop, zhou2022learning} is a parameter-efficient fine-tuning method, which keeps the VLM parameters fixed and only trains additional prompt tokens,
achieving comparable or superior performance to full fine-tuning while maintaining high efficiency~\cite{zhou2023zegclip, zhu2023segprompt, li2024cascade}.
%
%
Building on this paradigm, numerous studies \cite{yao2023visual, lee2023read, derakhshani2023bayesian, bulat2023lasp, khattak2023self, yao2024tcp, zheng2024large, kim2024aapl, zhang2024dept, roy2024consistency} have sought to enhance the adaptation performance of VLMs.
Specifically, Zhou \textit{et al.} \cite{zhou2022conditional} propose conditional context optimization by training a meta-network to generate image-condition prompt tokens, Yao \textit{et al.}~\cite{yao2023visual} introduce knowledge-guided context optimization, and Lee \textit{et al.} \cite{lee2023read} present a read-only prompt optimization framework by masked attention. 
Other notable contributions include text-to-text optimization strategy~\cite{bulat2023lasp}, self-regularization strategies~\cite{khattak2023self}, and class-aware prompt tuning approach~\cite{yao2024tcp}. Additionally, Chen \textit{et al.} integrate optimal transport into prompt tuning~\cite{chen2023plot}, and Zhang \textit{et al.} propose decoupled prompt tuning~\cite{zhang2024dept}. Roy \textit{et al}. emphasize consistency in the prompt tuning process to prevent overfitting~\cite{roy2024consistency}. 

{Moreover, recent studies}~\cite{fang2024aligning, bie2024xcoop, cao2024domain} {also explore integrating domain-specific knowledge into the prompt tuning process}, utilizing GPT-4~\cite{Gpt4} or pre-trained MedSAM~\cite{ma2024segment}.
{Additionally, recent studies}~\cite{shakeri2024few, koleilat2024biomedcoop} {have delved into investigating the potential of both generalist and specialist medical VLMs for few-shot medical image classification. 
For instance, Koleilat \textit{et al.}}~\cite{koleilat2024biomedcoop} {have expanded prompt tuning for few-shot medical image classification by incorporating large language models.}
{They achieve effective prompt tuning by leveraging semantic consistency with prompt ensembles derived from LLMs, and propose an effective knowledge distillation strategy to facilitate effective few-shot learning for novel biomedical classes.}
For a comprehensive overview of prompt tuning methods, readers are referred to recent surveys~\cite{gu2023systematic, xing2024survey}.

\vspace{-0.5em}
\subsection{{Representation Fine-Tuning}}

{Recently, Wu\textit{ et al.}}\cite{wu2024reft} {introduced the Representation Fine-Tuning (ReFT) paradigm as an alternative to parameter-efficient fine-tuning methods, such as prompt tuning}\cite{li2021prefix}, {LoRA}~\cite{hulora}, {and model adapters}~\cite{gao2024clip}.
{ReFT is motivated by the interchange intervention studies}~\cite{geiger2021causal} {that aim to interpret deep learning models by establishing the causal role of a representation or testing the concepts it encapsulates.
Specifically, ReFT freezes the model parameters and intervenes in the model representations to steer the model's behavior toward adaptation on the downstream task.
This process is operated in a linear subspace using the learnable intervention function, thus it is highly efficient.
In this paper, we are motivated by this adaptation technique, by utilizing it to incorporate medical domain knowledge into the prompts for the VLM, in order to construct disease-specific and instance-adaptive prompts to enhance the transferability of VLM on the medical image classification task.}

\section{Methodology}\label{sec:method}
In this section, we begin with a brief review of the vision-language foundation model and the prompt tuning paradigm. 
Subsequently, we introduce the overall pipeline of the proposed CILMP framework. 
We then provide a comprehensive explanation of the conditional intervention mechanism within the CILMP framework, which adaptively modulates the intervention on LLM representations based on the input image, thereby generating class-specific and instance-adaptive prompts. 
Finally, we summarize the training loss and the inference process of our method.

\begin{figure*}[thbp]
\centering
\includegraphics[width=1.0\textwidth]{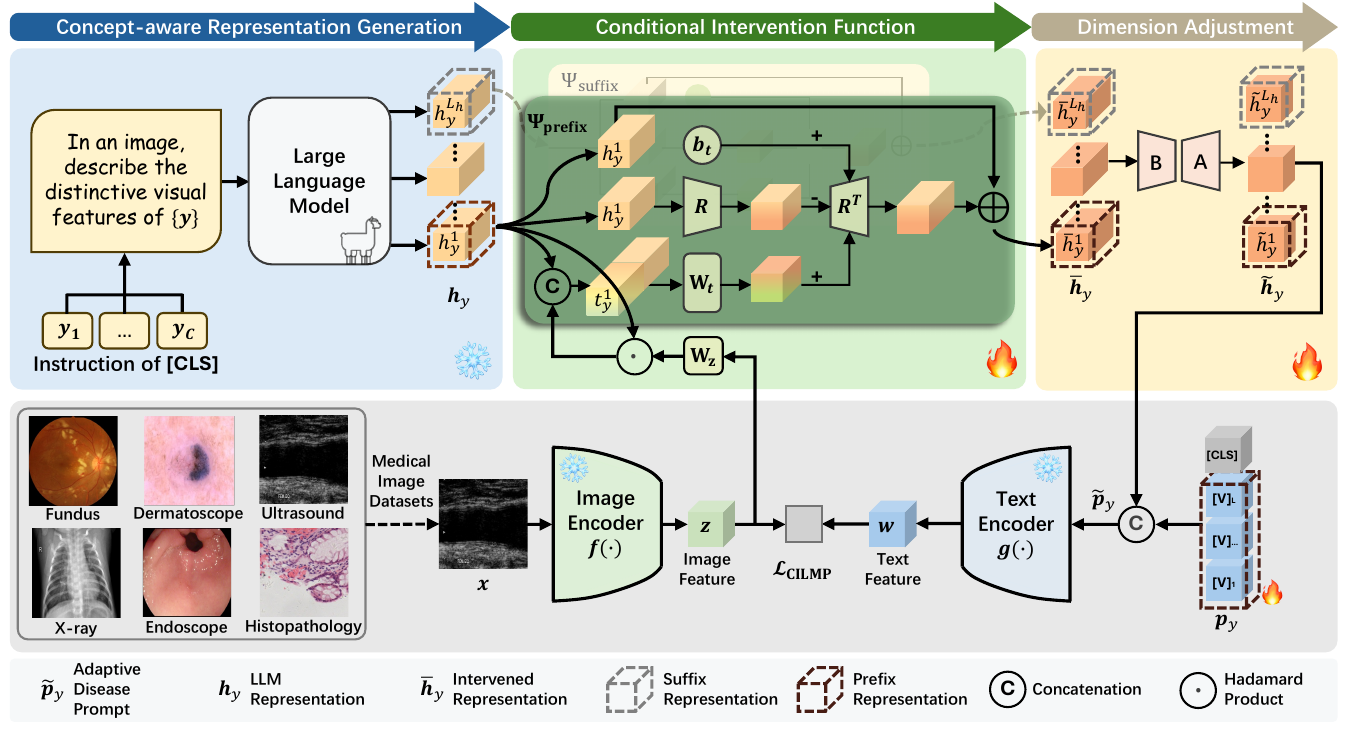}
\caption{{{Illustration of the CILMP framework.} CILMP first extracts concept-aware representations $\bm{h}_y$ from an LLM. Then, a conditional intervention function is introduced to adapt these representations towards accurate disease label prediction, producing intervened representations $\bar{\bm{h}}_y$.
After dimension adjustment, $\tilde{\bm{h}}_y$ are concatenated with the original prompts $\bm{p}_y$ to generate the adaptive disease prompts $\Tilde{\bm{p}}_y$.
Finally, $\mathcal{L}_{\text{CILMP}}$ is used to guide the prompt tuning process for the VLM.}}
\vspace{-0.5em}
\label{fig_cilmp}
\end{figure*}

\begin{figure}[thbp]
\includegraphics[width=0.5\textwidth]{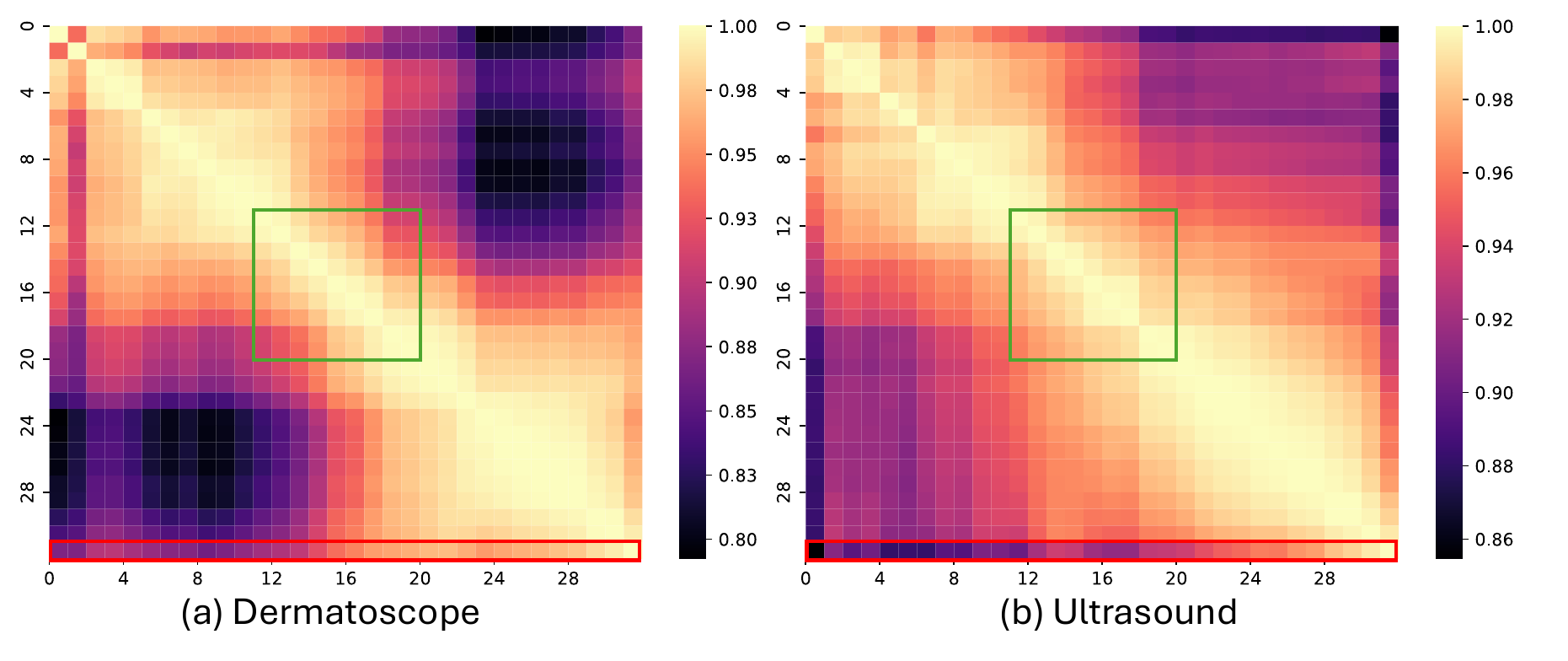}
\vspace{-1.5em}
\caption{{Centered kernel alignment heatmap}~\cite{kornblith2019similarity} between representations from different layers of the LLaMA3-8B~\cite{llama3modelcard}. 
The \textcolor{red}{red} box (last row) displays the similarity between representation from the last layer and those from other layers, while the \textcolor{mydarkgreen}{green} box highlights the similarity between adjacent layers.}
\label{fig_cka}
\vspace{-1em}
\end{figure}

\vspace{-0.5em}
\subsection{Preliminary}

\subsubsection{Vision-Language Foundation Model}

We first briefly introduce the Vision-Language Foundation Model, using the groundbreaking CLIP model \cite{radford2021learning} as an illustrative example.
CLIP consists of an image encoder $f(\cdot)$ and a text encoder $g(\cdot)$, both trained via contrastive learning \cite{oord2018representation} on a massive number of image-text pairs.
Typically, the image and text encoders are instantiated as a Vision Transformer \cite{dosovitskiy2020image} and a standard Language Transformer \cite{devlin2019bert}, respectively.
During training, CLIP considers each image $\bm{x}$ and its associated text $\bm{t}$ as a positive pair and utilizes the InfoNCE loss \cite{oord2018representation} for instance discrimination.
Specifically, CLIP first extracts the image and text embeddings via 
\begin{align}
        \bm{z} &= \text{Normalize}\left(f(\bm{x})\right), \\
        \bm{w} &= \text{Normalize}\left(g(\bm{t})\right),
\end{align}
where $\bm{z}$ is derived from a learnable classification token prepended to the image tokens and $\bm{w}$ is drawn from the $[EOS]$ token appended after the text description \cite{dosovitskiy2020image, devlin2019bert}.
The InfoNCE loss is then applied to these normalized embeddings for pre-training:
\begin{align}
    L_{\text{CLIP}} & = 0.5 \times (L_{v} + L_{t}), \\
    \text{where}\ \  L_{v} & = - \frac{1}{N} \sum_{i=1}^N \log \frac{\exp (\bm{z}_i^\mathrm{T} \bm{w}_{i} / \tau)}{ \sum_{j=1}^N \exp(\bm{z}_i^\mathrm{T} \bm{w}_{j} / \tau)}, \\
    L_{t} & = - \frac{1}{N} \sum_{i=1}^N \log \frac{\exp (\bm{w}_i^\mathrm{T} \bm{z}_{i} / \tau)}{ \sum_{j=1}^N \exp(\bm{w}_i^\mathrm{T} \bm{z}_{j} / \tau)},
\end{align}
where $N$ denotes the batch size during training and $\tau$ is a learnable temperature parameter used to scale the logits.

\subsubsection{Prompt Tuning}
%

Prompt tuning could adapt CLIP to specific downstream tasks without the need to fine-tune all the model parameters~\cite{zhou2022learning}.
Specifically, the prompt of each class is designed by
\begin{align}
    \bm{p} = [V]_1[V]_2[...][V]_L[CLS],
    \label{eq:prompt}
\end{align}
where each $[V]_l, (l \in \{1, ..., L\})$ is a learnable context vector,
$[CLS]$ is the embedding of each class name, and $L$ is a hyper-parameter specifying the number of learnable context tokens.
During fine-tuning, the parameters of the CLIP encoders are frozen, and only the context vectors are optimized using the 
contrastive training objective:
\begin{equation}
\centering
    L_{\text{prompt}} = - \frac{1}{N} \sum_{i=1}^N \log \frac{\exp (\bm{z}_i^\mathrm{T} \bm{w}_{{y}_i} / \tau)}{ \sum_{c=1}^C \exp(\bm{z}_i^\mathrm{T} \bm{w}_{c} / \tau)},
    \label{eq:prompt_tuning}
\end{equation}
where $\bm{y}_i$ is the class index of $\bm{z_i}$, $C$ is the total number of classes.
%

Although the existing prompt tuning paradigm has achieved remarkable performance with minimal training overhead, it exhibits two significant limitations. 
First, it lacks the specialized medical domain knowledge crucial for computer-aided diagnosis tasks. 
Second, the learnable context vectors are generally shared across categories, which prevents them from serving as discriminative features for specific diseases.
In this paper, we address these challenges by proposing a CILMP approach.

\subsection{The CILMP Framework}

Recent studies \cite{yang2023language, roy2024consistency, zheng2024large} have demonstrated that LLMs are effective candidates for providing concept-specific rich knowledge for various purposes.
Therefore, we propose the CILMP framework (Fig.~\ref{fig_cilmp}), which leverages the encyclopedic knowledge embedded in pre-trained large language models to address the limitations of existing prompt tuning methods.

\subsubsection{Concept-aware Representation Generation}
Given a dataset $\mathcal{D} = \{\bm{x}_i, {y}_i\}_{i=1}^{N}$, where $y \in \{1,...,C\}$ denotes the disease types corresponding to each image $\bm{x}$ in a specific modality (\textit{e.g.}, dermatoscopy images),
we generate concept-aware representations $\bm{h}_{y}$ by querying the LLM for class ${y}$ (\textit{e.g.} basal cell
carcinom). 
Specifically, we prompt the LLM with the question: ``In an image, describe the distinctive visual features of \{$y$\}".
Then LLM responds with a sequence representation for $y$, denoted by $\bm{h}_{y} \in \mathbb{R}^{L_h\times D_h}$, $L_h$ represents the sequence length and $D_h$ denotes the hidden dimensionality of the LLM.
Each element $\bm{h}_y^l$ ($l\in \{1, ..., L_h\}$) in $\bm{h}_y$ is drawn from the last token (\textit{i.e.}, {the EOS token}) of each independent layer of the LLM, as it encapsulates the information for the entire response.
{
Notably, these elements contain different semantic information.
To support this assertion, we compare the representations from different layers of the LLM using Centered Kernel Alignment}~\cite{kornblith2019similarity}{, a powerful tool for determining the correspondence between the hidden layers of neural networks.
As shown in Fig.}~\ref{fig_cka}{, the heatmap reveals a substantial dissimilarity between the last layer and the preceding layers, as well as noticeable differences between adjacent layers.
This underscores the importance of utilizing these elements, each containing distinct semantic information, to enhance the process of concept-aware representation generation.}
As such, we derive $\bm{h}_y$, a concept-aware representation that encapsulates rich medical knowledge about disease $y$, because it implies the LLM's comprehensive understanding of the queried disease type.

%

\subsubsection{Adaptive Disease Prompts Construction}

The inherent domain gap between the representation spaces of LLMs and VLMs hinders direct knowledge transfer between them. To address this, we introduce an additional adaptation function, a meta-function $\Psi_{\theta}(\cdot)$ parameterized by $\theta$. $\Psi_{\theta}(\cdot)$ maps $\bm{h}_y$ into the VLM feature space by optimizing towards the true disease label.
Formally, the adapted LLM representation $\overline{\bm{h}}_y$ for each class $y$ is obtained via
\begin{align}
    \overline{\bm{h}}_y = \Psi_{\theta}(\bm{h}_y).
\end{align}

Then we concatenate $\overline{\bm{h}}_y$ with the original prompts $\bm{p}_y$ (obtained via Eq. (\ref{eq:prompt})) to form the adaptive disease prompt for the text encoder.
%
To deal with the dimension mismatch problem between $\overline{\bm{h}}_y$ and $\bm{p}_y$, we introduce a linear projection layer $\bm{W}$ to adjust the dimensions.
%
%
%
To maintain parameter efficiency, we employ a low-rank decomposition \cite{hulora} for $\bm{W}$, {generating $ \widetilde{\bm{h}}_y \in \mathbb{R}^{L_h \times D_p}$}:
\begin{align}
    \widetilde{\bm{h}}_y =\bm{W}\overline{\bm{h}}_y =   \bm{B}\bm{A}\overline{\bm{h}}_y,
\end{align}
where $\bm{A} \in \mathbb{R}^{r\times D_h}$ and $\bm{B} \in \mathbb{R}^{D_p \times r}$. 
Here, $r$ denotes the low-rank dimensionality and $D_p$ represents the dimensionality of original prompt tokens.
Finally, following \cite{zheng2024large}, the adaptive disease prompts for $y$, {denoted by $\widetilde{\bm{p}}_y \in \mathbb{R}^{(L_h + L)\times D_p}$}, is defined as
\begin{align}
    \widetilde{\bm{p}}_y := \text{concat}[\widetilde{\bm{h}}_y,\  \bm{p}_y],
    \label{eq:knowledge_enhanced}
\end{align}

\subsection{Conditional Intervention Mechanism}

\subsubsection{Unconditional Intervention}
The goal of CILMP is to control the transformation of LLM representations through targeted interventions, directing the entire pipeline toward predicting disease labels.
To this end, an unconditional intervention mechanism \cite{wu2024reft} can be introduced.
Specifically, given $\bm{h}_y^{l} \in \mathbb{R}^{D_h}$ at the $l$-th ($l\in\{1, ..., L_h\}$) position of the sequential LLM representation $\bm{h}_y$, the adaptation process can be formulated as a standard interchange-based intervention function \cite{wu2024reft} applied on top of $\bm{h}^{l}_y$.
Formally, $\Psi: \mathbb{R}^{D_h} \rightarrow \mathbb{R}^{D_h}$ is defined as 
\begin{align}
    \Psi_{\theta}(\bm{h}_y^l) = \bm{h}^l_y + \bm{\mathrm{R}}^\mathrm{T} (\bm{\mathrm{W}}_h\bm{h}_y^l + \bm{\mathrm{b}}_h - \bm{\mathrm{R}}\bm{h}_y^l),
    \label{eq:11}
\end{align}
where $\bm{\mathrm{R}} \in \mathbb{R}^{r\times D_h}$ is a low-rank projection matrix, $\bm{\mathrm{W}}_h \in \mathbb{R}^{r\times D_h}$, $\bm{\mathrm{b}}_h \in \mathbb{R}^{r}$, and $r$ is the dimensionality of the subspace being intervened upon.
The learnable parameters are thus composed as $\theta = \{\bm{\mathrm{R}}, \bm{\mathrm{W}}_h, \bm{\mathrm{b}}_h \}$.

In this formulation, $\bm{\mathrm{W}}_h\bm{h}^l_y + \bm{\mathrm{b}}_h$ represents a learned projected term that modifies the original $\bm{h}^l_y$ within a linear subspace spanned by the rows of $\bm{\mathrm{R}}$.
{
In other words, during supervision, Eq.}~\ref{eq:11} {learns to ``edit" the LLM representation by embedding the concept of the target disease label into it.
The editing process operates within a linear subspace, where  $\bm{\mathrm{R}}$ is trained to project the original LLM representation into this subspace.
Therefore, this can be explained as finding the subspace that maximizes the probability of the desired output after the intervention. 
Notably, this process is highly efficient, as the learnable parameters are represented by low-rank matrices.}



%
\subsubsection{Conditional Intervention}

The aforementioned process utilizes an unconstrained intervention mechanism, which means the learned projected term $\bm{\mathrm{W}}_h\bm{h}^l_y + \bm{\mathrm{b}}_h$ remains unrestricted.
{However, we seek to enable the generation of adaptive prompts that not only incorporate medical knowledge specific to each category, but are also tailored to each individual image,  in order to enhance the flexibility and adaptability}~\cite{zhou2022conditional, derakhshani2023bayesian}.
As such, we further incorporate the matching prior~\cite{zhou2023zegclip} into the intervention function.
{The matching prior quantifies modality compatibility between image and text representations before the final cross-modal alignment, which aims at preconditioning the adaptive prompt generation on each individual image example.}
To achieve this, we further propose a conditional intervention mechanism, which conditions the intervention process on a Relationship Descriptor (RD) between the image representation $\bm{z}$ and each {non-intervened} representation $\bm{h}^l_y$.
Following \cite{zhou2023zegclip, li2024cascade}, the RD, denoted by $\bm{t}_y^l\in \mathbb{R}^{2D_h}$, is calculated by 
\begin{align}
    \bm{t}_y^l = \text{concat}[\bm{h}^l_y, \ \bm{h}^l_y \odot \bm{\mathrm{W}}_z\bm{z}],
\end{align}
where $\odot$ is the Hadamard product, $\bm{\mathrm{W}}_z \in \mathbb{R}^{D_h \times D_p}$ is introduced for size adjustment.
%
%
To maintain parameter efficiency, we also decompose $\bm{\mathrm{W}}_z$ with two low-rank matrices~\cite{hulora}.
Subsequently, we leverage RD to intervene on the LLM representation within the $r$-dimensional linear subspace, as given by
\begin{align}
    \Psi_{\theta}(\bm{h}_y^l, \bm{z}) = \bm{h}^l_y + \bm{\mathrm{R}}^\mathrm{T} (\bm{\mathrm{W}}_t\bm{t}_y^l + \bm{\mathrm{b}}_t - \bm{\mathrm{R}}\bm{h}_y^l),
    \label{eq:13}
\end{align}
where $\bm{\mathrm{W}}_t \in \mathbb{R}^{r\times 2D_h}$ and $\bm{\mathrm{b}}_t \in \mathbb{R}^{r}$.
The set of learnable parameters is therefore updated to $\theta = \{\bm{\mathrm{R}}, \bm{\mathrm{W}}_t,  \bm{\mathrm{W}}_z, \bm{\mathrm{b}}_t\}$.
%


{In addition, given that the large language model generates a sequence of representations, addressing how to intervene effectively across the entire sequence presents another problem.
Unlike PEFT methods, representation fine-tuning operates on a sequence of representations, necessitating interventions on each element in the sequence.
However, intervening on all elements can lead to excessive computational burden, especially when dealing with lengthy sequences. 
This excessive intervention may impede extendability and risk the loss of valuable medical knowledge embedded in the original LLM representation}~\cite{wu2024reft}.
{To strike a balance between adaptation and knowledge retention, a compromise strategy should be devised to determine efficient interventions across the entire sequence. 
Drawing inspiration from ReFT, a practical approach involves intervening on a subset of them, where the length of intervention acts as a hyperparameter. 
This not only maintains efficiency but also introduces flexibility.}
In light of this, our CILMP framework adopts a simple bilateral intervention strategy~\cite{wu2024reft}.
Specifically, we propose the development of separate intervention functions for the prefix and suffix segments of the LLM representations.
In this context, the intervention can be encapsulated as $\langle\Psi, P\rangle$, where $P$ denotes the set of positions within the LLM representation at which $\Psi$ is applied.
Hence, the prefix intervention $\langle\Psi_{\text{prefix}}, P_{\text{prefix}}\rangle$ targets positions $P_{\text{prefix}} = \{1, ..., L_{\text{prefix}}\}$, while the suffix intervention $\langle\Psi_{\text{suffix}}, P_{\text{suffix}}\rangle$ focuses on positions $P_{\text{suffix}} = \{L_h - L_{\text{suffix}}, ..., L_h \}$, where $L_{\text{prefix}}$ and $L_{\text{suffix}}$ are hyper-parameters that specify the intervention lengths.

\subsection{Model Optimization and Inference}
After obtaining the adaptive disease prompts $\widetilde{\bm{p}}$ via Eq. (\ref{eq:knowledge_enhanced}) for each category, we proceed by freezing the parameters of the VLM.
Then, we utilize the supervised contrastive loss \cite{khosla2020supervised} to guide the prompt tuning process as
\begin{align}
    \mathcal{L}_{\text{CILMP}} =  - \frac{1}{N} \sum_{i=1}^N \log \frac{\exp (\bm{z}_i^\mathrm{T} g(\widetilde{\bm{p}}_{y_i}) / \tau)}{ \sum_{c=1}^C \exp(\bm{z}_i^\mathrm{T} g(\widetilde{\bm{p}}_{c}) / \tau)},
\end{align}
where $N$ denotes the number of samples in a training batch, and $\tau$ is the learned temperature parameter.

During inference, the CILMP framework predicts the disease label for a given image $\bm{x}$ by comparing the cosine similarity between the image embedding and the embeddings of each enhanced prompt.
In particular, the probability that the image $\bm{x}$ belongs to class $c$ is computed by
\begin{equation}
        p(y=c|\bm{x}) = \frac{\exp (f(\bm{x})^\mathrm{T} g(\widetilde{\bm{p}}_{c}) / \tau)}{ \sum_{j=1}^C \exp(f(\bm{x})^\mathrm{T} g(\widetilde{\bm{p}}_{j}) / \tau)}.
\end{equation}

\section{Experiments}\label{sec:exp}
\subsection{Datasets}
There are 11 datasets utilized for evaluation, which encompass six different data modalities.
Specifically, we conduct experiments on the dermatoscope modality using the DermaMNIST \cite{medmnistv2}, Derm7pt \cite{kawahara2019skin}, and ISIC 2018 \cite{ISIC} datasets. 
For the fundus modality, we use the ADAM \cite{ADAM}, APTOS 2019 \cite{APTOS2019}, and ODIR \cite{ODIR} datasets.
Additionally, ultrasound images are evaluated using the Fetal-US \cite{Fetal} dataset, histopathology images are evaluated with the Chaoyang \cite{Chaoyang} dataset, and endoscope images are assessed using the Kvasir \cite{Kvasir} dataset. 
X-ray images are incorporated using the CPN-X-ray \cite{shastri2022cheximagenet} and the Pneumonia \cite{kermany2018identifying} datasets.
TABLE \ref{tab:statistic} summarizes the number of classes, training images, and test images for each dataset employed in this paper.

\begin{table}[t]
\centering
\renewcommand\arraystretch{0.93}
\caption{The statistics of 11 datasets over 6 modalities.
\label{tab:statistic}
}
\resizebox{0.45\textwidth}{!}{
\begin{tabular}{l|l|c|l|l}
\toprule
\textbf{Dataset}     & \textbf{Modality} & \textbf{\# Classes}      & \textbf{\# Training} & \textbf{\# Test}  \\ \midrule
ADAM \cite{ADAM}       & Fundus   & 2       & 400         & 400            \\
APTOS 2019 \cite{APTOS2019}  & Fundus    & 5        & 2,930       & 732          \\
Chaoyang \cite{Chaoyang}    & 
Histopathology & 4   & 4,021       & 2,139        \\
CPN-X-ray \cite{shastri2022cheximagenet}    & 
X-ray & 3   &  3,659    & 1,569     \\
DermaMNIST \cite{medmnistv2}  & Dermatoscope  & 7     & 5,600       & 2,400        \\
Derm7pt \cite{kawahara2019skin}         & Dermatoscope    & 5        & 413       & 395        \\
Fetal-US \cite{Fetal}    & Ultrasound  & 6      & 7,129       & 5,271        \\
ISIC 2018 \cite{ISIC}   & Dermatoscope  & 7    & 10,015      & 1,512        \\
Kvasir \cite{Kvasir}      & Endoscope   & 8     & 5,600       & 2,400         \\
ODIR \cite{ODIR}        & Fundus    & 8        & 5,113       & 1,279        \\
Pneumonia \cite{kermany2018identifying} & X-ray   & 3       & 5,216       & 624            \\
\bottomrule
\end{tabular}}
\vspace{-1em}
\label{tab-data}
\end{table}

\begin{table*}[t]
\centering
\renewcommand\arraystretch{0.93} \setlength\tabcolsep{6pt}
\caption{{Comparison with state-of-the-art methods on 5 datasets including ADAM, APTOS 2019, Chaoyang, CPN-X-ray, and DermaMNIST}. The best and the second best results are highlighted in \best{red} and \second{blue}, \textcolor{black}{respectively.} \textcolor{black}{``Param." is the number of trainable parameters.}}
\label{tab-sota-results1}
\resizebox{0.49\textwidth}{!}{
\begin{subtable}{.53\linewidth} \centering
\setlength\tabcolsep{4.5pt}
\begin{tabular}{l|c|cccc}
\toprule
\textbf{Method   }  &  \textbf{Param.}         & \textbf{ACC} [\%]   & \textbf{F1} [\%]    & \textbf{AUC} [\%]   & \textbf{Kappa} [\%] \\ \midrule
CoOp \cite{zhou2022learning}& $2$K & $77.0$ & $63.5$ & $91.6$ & $61.1$ \\
CoCoOp \cite{zhou2022conditional} & $35$K & $78.1$ & $65.7$ & $91.5$ & $62.3$ \\
KgCoOp \cite{yao2023visual}& $2$K & $68.7$ & $47.6$ & $84.6$ & $42.7$ \\
CLIP-Adapter \cite{gao2024clip}& $131$K & $74.9$ & $57.9$ & $89.0$ & $56.9$ \\
RPO \cite{lee2023read}& $5$K & $81.6$ & $71.5$ & $93.9$ & $69.2$ \\
LASP \cite{bulat2023lasp}& $39$K & $81.1$ & $69.9$ & $92.9$ & $68.1$ \\
PromptSRC \cite{khattak2023self}& $46$K & $81.8$ & $69.6$ & $93.9$ & $69.1$ \\
TCP \cite{yao2024tcp}& $332$K & $71.1$ & $50.8$ & $86.9$ & $47.6$ \\
AAPL \cite{kim2024aapl}& $35$K & $77.8$ & $65.3$ & $91.2$ & $62.1$ \\
PLOT++ \cite{chen2023plot}& $14$K & $81.9$ & $72.4$ & $94.1$ & $69.9$ \\
 {DCPL}  \cite{cao2024domain}& $3.8$M & \second{$84.5$}  & \second{$76.7$} & \second{$95.1$} &  \second{$74.1$}\\  {ViP} \cite{fang2024aligning}& $41.5$M & $78.8$ & $65.9$ & $91.2$ & $63.1$ \\  {XCoOp} \cite{bie2024xcoop}& $94$K & $82.6$  & $74.3$  & $94.4$ & $70.8$ \\ DePT \cite{zhang2024dept}& $48$K & {$83.2$} & {$72.5$} & {$94.6$} & {$72.2$} \\
CoPrompt \cite{roy2024consistency}&  $4.7$M & $83.0$ & $74.2$ & {$94.7$} & $71.7$ \\ 
\midrule
\textbf{CILMP (Ours)}& $3.8$M & \best{$86.2$} & \best{$79.1$} & \best{$96.0$} & \best{$77.2$} \\
\bottomrule
\end{tabular}
 \caption{\textbf{Average over 11 datasets}\label{tab:average}}
\end{subtable}\hfill}
\resizebox{0.49\textwidth}{!}{
\begin{subtable}{.53\linewidth} \centering
\begin{tabular}{l|cccc}
\toprule
\textbf{Method   }           & \textbf{ACC} [\%]   & \textbf{F1} [\%]    & \textbf{AUC} [\%]   & \textbf{Kappa} [\%] \\ \midrule
CoOp \cite{zhou2022learning} & $84.0\pm1.5$ & $77.5\pm2.4$ & $86.3\pm0.8$ & $55.0\pm4.7$ \\
CoCoOp \cite{zhou2022conditional} & $81.5\pm0.6$ & $74.3\pm0.9$ & $84.3\pm1.6$ & $48.6\pm1.7$ \\
KgCoOp \cite{yao2023visual} & $79.8\pm0.0$ & $56.3\pm1.2$ & $81.5\pm1.0$ & $18.2\pm1.6$ \\
CLIP-Adapter \cite{gao2024clip} & $82.4\pm1.0$ & $74.9\pm1.1$ & $84.8\pm0.5$ & $49.8\pm2.2$ \\
RPO \cite{lee2023read} & $87.8\pm1.2$ & $82.7\pm1.6$ & $92.7\pm0.4$ & $65.5\pm3.2$ \\
LASP \cite{bulat2023lasp} & $86.8\pm0.6$ & $81.3\pm0.6$ & $90.5\pm0.6$ & $62.6\pm1.2$ \\
PromptSRC \cite{khattak2023self} & \second{$88.2\pm0.3$} & $83.2\pm1.2$ & $91.8\pm1.6$ & $66.5\pm2.3$ \\
TCP \cite{yao2024tcp} & $80.5\pm0.0$ & $60.5\pm0.6$ & $83.4\pm0.3$ & $24.7\pm0.9$ \\
AAPL \cite{kim2024aapl} & $82.0\pm0.9$ & $75.6\pm0.9$ & $85.3\pm0.6$ & $51.4\pm1.8$ \\
PLOT++ \cite{chen2023plot} & $87.6\pm1.2$ & $82.4\pm1.4$ & $92.3\pm0.2$ & $64.8\pm2.9$ \\
 {DCPL}  \cite{cao2024domain} &  {$86.3\pm0.8$} & $80.7\pm0.9$ & $89.8\pm0.5$ & $61.4\pm1.8$ 
\\ {ViP} \cite{fang2024aligning}  & 
$83.9\pm0.1$ & $72.6\pm0.4$ & $83.9\pm1.0$&  $45.9\pm0.8$
\\  XCoOp \cite{bie2024xcoop} & $ 84.7 \pm 0.5$ &  $ 79.6 \pm 0.6$ &  $ 89.1 \pm 0.1$ &  $ 59.3 \pm 1.3$
 \\ DePT \cite{zhang2024dept} & {$88.1\pm0.9$} & \second{$83.9\pm1.1$} & \second{$93.0\pm0.2$} & \second{$70.0\pm2.1$} \\
CoPrompt \cite{roy2024consistency} & $87.6\pm0.2$ & $82.3\pm0.4$ & $90.7\pm0.6$ & $64.6\pm0.9$ \\

\midrule
\textbf{CILMP (Ours)} & \best{$90.1\pm0.4$} & \best{$85.4\pm0.6$} & \best{$93.5\pm0.3$} &\best{$70.9\pm1.2$}  \\ \bottomrule
\end{tabular}
 \caption{ADAM}
\end{subtable}}
\vfill

\resizebox{0.49\textwidth}{!}{
\begin{subtable}{.53\linewidth} \centering
\begin{tabular}{l|cccc}
\toprule
\textbf{Method   }           & \textbf{ACC} [\%]   & \textbf{F1} [\%]    & \textbf{AUC} [\%]   & \textbf{Kappa} [\%] \\ \midrule
CoOp \cite{zhou2022learning} & $76.8\pm0.2$ & $40.2\pm0.6$ & $91.1\pm0.2$ & $62.3\pm0.2$ \\
CoCoOp \cite{zhou2022conditional} & $79.2\pm0.6$ & $51.8\pm1.4$ & $92.2\pm0.1$ & $66.8\pm0.9$ \\
KgCoOp \cite{yao2023visual} & $74.1\pm0.3$ & $32.7\pm0.1$ & $87.0\pm0.4$ & $57.8\pm0.4$ \\
CLIP-Adapter \cite{gao2024clip} & $73.2\pm0.1$ & $32.3\pm0.1$ & $84.2\pm0.1$ & $56.3\pm0.2$ \\
RPO \cite{lee2023read} & $81.9\pm0.4$ & $61.4\pm1.4$ & $93.6\pm0.1$ & $71.7\pm0.7$ \\
LASP \cite{bulat2023lasp} & $76.0\pm0.2$ & $38.9\pm0.7$ & $88.7\pm0.2$ & $60.8\pm0.3$ \\
PromptSRC \cite{khattak2023self} & $76.8\pm0.6$ & $38.3\pm2.3$ & $92.7\pm0.4$ & $62.3\pm1.0$ \\
TCP \cite{yao2024tcp} & $74.4\pm0.2$ & $32.7\pm0.1$ & $87.9\pm0.1$ & $58.2\pm0.2$ \\
AAPL \cite{kim2024aapl} & $78.4\pm1.0$ & $48.9\pm3.6$ & $91.8\pm0.2$ & $65.4\pm1.7$ \\
PLOT++ \cite{chen2023plot} & {$83.5\pm0.4$} & $65.5\pm0.3$ & {$94.3\pm0.1$} & {$74.5\pm0.6$} \\
{DCPL}  \cite{cao2024domain}  & \second{$83.6\pm0.3$} & $64.7\pm0.1$ & \second{$94.4\pm0.1$} & \second{$74.7\pm0.4$}\\ 
ViP \cite{fang2024aligning} & $78.8\pm0.5$ & $50.9\pm1.4$ & $90.9\pm0.4$ & $66.3\pm0.8$  \\  XCoOp \cite{bie2024xcoop} & $ 80.9 \pm 0.6$ &  $ 59.6 \pm 1.2$ &  $ 93.3 \pm 0.1$ &  $ 70.1 \pm 0.9$
 \\ DePT \cite{zhang2024dept} & $81.7\pm0.4$ & $56.0\pm2.2$ & $94.1\pm0.2$ & $71.0\pm0.8$ \\
CoPrompt \cite{roy2024consistency} & $82.7\pm0.9$ & \second{$66.0\pm1.8$} & $94.2\pm0.2$ & $73.3\pm1.5$ \\ 
\midrule

\textbf{CILMP (Ours)} & \best{$84.5\pm0.5$} & \best{$66.7\pm0.8$} & \best{$94.6\pm0.2$} &\best{$75.7\pm0.7$} \\
\bottomrule
\end{tabular}
 \caption{APTOS 2019}
\end{subtable}\hfill}
\resizebox{0.49\textwidth}{!}{
\begin{subtable}{.53\linewidth} \centering
\begin{tabular}{l|cccc}
\toprule
\textbf{Method   }           & \textbf{ACC} [\%]   & \textbf{F1} [\%]    & \textbf{AUC} [\%]   & \textbf{Kappa} [\%] \\ \midrule
CoOp \cite{zhou2022learning} & $74.1\pm0.4$ & $65.5\pm0.3$ & $89.4\pm0.2$ & $62.3\pm0.6$ \\
CoCoOp \cite{zhou2022conditional} & $76.5\pm0.4$ & $68.5\pm0.4$ & $90.6\pm0.2$ & $65.8\pm0.4$ \\
KgCoOp \cite{yao2023visual} & $65.0\pm0.5$ & $55.6\pm0.3$ & $83.4\pm0.2$ & $48.2\pm0.6$ \\
CLIP-Adapter \cite{gao2024clip} & $73.0\pm0.3$ & $63.6\pm0.3$ & $88.7\pm0.2$ & $60.4\pm0.5$ \\
RPO \cite{lee2023read} & $78.4\pm0.6$ & $71.0\pm0.2$ & $91.4\pm0.4$ & $68.6\pm0.9$ \\
LASP \cite{bulat2023lasp} & $78.8\pm0.3$ & $71.4\pm0.3$ & $92.1\pm0.3$ & $69.3\pm0.4$ \\
PromptSRC \cite{khattak2023self} & $78.8\pm1.0$ & $69.8\pm1.2$ & $92.9\pm0.4$ & $69.0\pm1.5$ \\
TCP \cite{yao2024tcp} & $67.6\pm0.4$ & $58.1\pm0.6$ & $85.4\pm0.2$ & $52.0\pm0.5$ \\
AAPL \cite{kim2024aapl} & $75.3\pm0.4$ & $67.0\pm0.5$ & $89.9\pm0.3$ & $64.1\pm0.4$ \\
PLOT++ \cite{chen2023plot} & $79.9\pm0.1$ & $73.7\pm0.1$ & $92.7\pm0.1$ & $71.0\pm0.2$ \\
{DCPL} \cite{cao2024domain} & \best{$82.3\pm0.3$} & \second{$76.0\pm0.3$} & \second{$94.0\pm0.1$} & \best{$74.5\pm0.5$}
\\ ViP \cite{fang2024aligning}   & $76.2\pm0.3$ & $66.8\pm1.4$ & $89.7\pm0.2$ & $65.2\pm0.4$ \\ 
 XCoOp \cite{bie2024xcoop} & $ 79.7 \pm 0.1$ &  $ 72.6 \pm 0.2$ &  $ 93.0 \pm 0.1$ &  $ 70.7 \pm 0.2$
 \\ DePT \cite{zhang2024dept} & $79.6\pm0.7$ & $71.4\pm0.4$ & $93.0\pm0.1$ & $70.3\pm1.1$ \\
CoPrompt \cite{roy2024consistency} & \second{$80.8\pm0.2$} & {$74.8\pm0.4$} & {$93.8\pm0.1$} & {$72.4\pm0.3$} \\ 
\midrule

\textbf{CILMP (Ours)} & \best{$82.3\pm0.3$} & \best{$76.1\pm0.5$} & \best{$94.1\pm0.1$} & \second{$74.4\pm0.4$} \\
\bottomrule
\end{tabular}
 \caption{Chaoyang}
\end{subtable}\hfill}
\resizebox{0.49\textwidth}{!}{
\begin{subtable}{.53\linewidth} \centering
\begin{tabular}{l|cccc}
\toprule
\textbf{Method   }           & \textbf{ACC} [\%]   & \textbf{F1} [\%]    & \textbf{AUC} [\%]   & \textbf{Kappa} [\%] \\ \midrule
CoOp \cite{zhou2022learning} & $93.7\pm0.3$ & $93.9\pm0.3$ & $99.0\pm0.1$ & $90.6\pm0.5$ \\
CoCoOp \cite{zhou2022conditional} & $94.9\pm0.1$ & $95.0\pm0.1$ & $99.2\pm0.1$ & $92.2\pm0.1$ \\
KgCoOp \cite{yao2023visual} & $83.7\pm0.5$ & $84.0\pm0.5$ & $96.4\pm0.0$ & $75.3\pm0.8$ \\
CLIP-Adapter \cite{gao2024clip} & $92.2\pm0.2$ & $92.3\pm0.1$ & $98.7\pm0.0$ & $88.2\pm0.2$ \\
RPO \cite{lee2023read} & \second{$96.8\pm0.1$} & \second{$96.9\pm0.1$} & \second{$99.6\pm0.0$} & \second{$95.1\pm0.2$} \\
LASP \cite{bulat2023lasp} & $95.5\pm0.2$ & $95.6\pm0.2$ & $99.4\pm0.0$ & $93.2\pm0.3$ \\
PromptSRC \cite{khattak2023self} & $92.7\pm0.1$ & $92.8\pm0.1$ & $99.2\pm0.1$ & $89.0\pm0.1$ \\
TCP \cite{yao2024tcp} & $86.4\pm0.2$ & $86.7\pm0.2$ & $97.3\pm0.1$ & $79.5\pm0.4$ \\
AAPL \cite{kim2024aapl} & $94.5\pm0.4$ & $94.6\pm0.3$ & $99.2\pm0.0$ & $91.7\pm0.5$ \\
PLOT++ \cite{chen2023plot} & $95.5\pm0.0$ & $95.7\pm0.0$ & $99.4\pm0.1$ & $93.3\pm0.1$ \\
{DCPL}  \cite{cao2024domain}  & $96.6\pm0.1$ & $96.7\pm0.1$ & \second{$99.6\pm0.0$} & $94.9\pm0.1$ \\ {ViP} \cite{fang2024aligning} & $94.6\pm0.2$ & $94.7\pm0.2$ & $99.0\pm0.1$ & $91.8\pm0.3$  \\  XCoOp \cite{bie2024xcoop} & $ 96.0 \pm 0.1$ &  $ 96.2 \pm 0.1$ &  \second{$ 99.6 \pm 0.0$} &  $ 94.1 \pm 0.1$
 \\ DePT \cite{zhang2024dept} & $95.2\pm0.7$ & $95.4\pm0.7$ & $99.3\pm0.0$ & $92.9\pm1.0$ \\
CoPrompt \cite{roy2024consistency} & \best{$97.7\pm0.1$} & \best{$97.8\pm0.0$} & \best{$99.8\pm0.0$} & \best{$96.5\pm0.1$} \\ 
\midrule

\textbf{CILMP (Ours)} & {$96.2\pm0.3$} & {$96.3\pm0.3$} & \second{$99.6\pm0.0$} & {$94.3\pm0.5$} \\
\bottomrule
\end{tabular}
 \caption{CPN-X-ray}
\end{subtable}\hfill}
\resizebox{0.49\textwidth}{!}{
\begin{subtable}{.53\linewidth} \centering
\begin{tabular}{l|cccc}
\toprule
\textbf{Method   }           & \textbf{ACC} [\%]   & \textbf{F1} [\%]    & \textbf{AUC} [\%]   & \textbf{Kappa} [\%] \\ \midrule
CoOp \cite{zhou2022learning} & $77.4\pm0.6$ & $50.5\pm1.3$ & $93.3\pm0.5$ & $55.8\pm1.3$ \\
CoCoOp \cite{zhou2022conditional} & $79.1\pm0.6$ & $57.6\pm1.8$ & $94.6\pm0.3$ & $60.6\pm0.9$ \\
KgCoOp \cite{yao2023visual} & $70.7\pm0.1$ & $26.4\pm0.3$ & $83.3\pm1.1$ & $32.8\pm0.8$ \\
CLIP-Adapter \cite{gao2024clip} & $76.7\pm0.2$ & $43.0\pm0.2$ & $90.2\pm0.2$ & $53.9\pm0.3$ \\
RPO \cite{lee2023read} & $82.0\pm0.6$ & $60.8\pm3.1$ & $95.6\pm0.1$ & $65.2\pm1.4$ \\
LASP \cite{bulat2023lasp} & $82.4\pm0.2$ & $65.9\pm0.5$ & $95.8\pm0.2$ & $66.4\pm0.3$ \\
PromptSRC \cite{khattak2023self} & $83.8\pm0.4$ & $64.1\pm2.0$ & $95.5\pm0.5$ & $68.1\pm0.8$ \\
TCP \cite{yao2024tcp} & $72.2\pm0.2$ & $29.2\pm1.0$ & $87.1\pm0.2$ & $37.3\pm0.5$ \\
AAPL \cite{kim2024aapl} & $78.9\pm0.1$ & $54.0\pm1.1$ & $93.4\pm0.2$ & $57.7\pm0.1$ \\
PLOT++ \cite{chen2023plot} & $82.9\pm0.3$ & $66.6\pm1.3$ & $95.9\pm0.1$ & $67.4\pm0.6$ \\
{DCPL}  \cite{cao2024domain} 
& \second{$86.8\pm0.5$} & \second{$75.9\pm0.6$} & \second{$97.9\pm0.1$} & \second{$74.8\pm0.7$}
\\ ViP \cite{fang2024aligning}  & $80.0\pm0.5$ & $55.1\pm3.1$ & $94.4\pm0.1$ & $59.9\pm0.8$ \\  XCoOp \cite{bie2024xcoop} & $ 84.3 \pm 0.1$ &  $ 71.1 \pm 0.7$ &  $ 96.7 \pm 0.1$ &  $ 69.9 \pm 0.3$
 \\ DePT \cite{zhang2024dept} & $83.8\pm0.6$ & $64.4\pm0.3$ & $95.9\pm0.1$ & $68.1\pm0.9$ \\
CoPrompt \cite{roy2024consistency} & {$84.6\pm0.4$} & {$69.3\pm0.3$} & {$97.0\pm0.1$} & {$70.0\pm0.5$} \\ 
\midrule

\textbf{CILMP (Ours)} & \best{$87.4\pm0.3$} & \best{$77.7\pm1.0$} & \best{$98.0\pm0.0$} & \best{$76.0\pm0.4$} \\
\bottomrule
\end{tabular}
 \caption{DermaMNIST}
\end{subtable}\hfill}
\vspace{-2.5em}
\end{table*}

\begin{table*}[t]
\centering
\renewcommand\arraystretch{0.93} \setlength\tabcolsep{6pt}
\caption{{Comparison with state-of-the-art methods on 6 datasets including Derm7pt, Fetal-US, ISIC 2018, Kvasir, ODIR, and Pneumonia}. The best and the second best results are highlighted in \best{red} and \second{blue} \textcolor{black}{, respectively.}} 
\label{tab-sota-results2}
\resizebox{0.49\textwidth}{!}{
\begin{subtable}{.53\linewidth} \centering
\begin{tabular}{l|cccc}
\toprule
\textbf{Method   }           & \textbf{ACC} [\%]   & \textbf{F1} [\%]    & \textbf{AUC} [\%]   & \textbf{Kappa} [\%] \\ \midrule
CoOp \cite{zhou2022learning} & $67.2\pm1.8$ & $43.9\pm5.1$ & $86.8\pm0.9$ & $41.3\pm2.9$ \\
CoCoOp \cite{zhou2022conditional} & $64.6\pm1.6$ & $33.4\pm2.5$ & $81.6\pm1.3$ & $31.5\pm3.5$ \\
KgCoOp \cite{yao2023visual} & $63.9\pm0.3$ & $35.1\pm0.8$ & $80.3\pm1.2$ & $28.9\pm0.8$ \\
CLIP-Adapter \cite{gao2024clip} & $69.5\pm0.1$ & $46.6\pm0.3$ & $84.8\pm0.2$ & $44.9\pm0.4$ \\
RPO \cite{lee2023read} & $70.7\pm0.9$ & $48.9\pm2.9$ & $87.9\pm0.6$ & $47.2\pm2.1$ \\
LASP \cite{bulat2023lasp} & $69.6\pm1.4$ & $49.4\pm2.0$ & $83.6\pm0.7$ & $44.3\pm3.6$ \\
PromptSRC \cite{khattak2023self} & $70.1\pm0.3$ & $46.5\pm1.9$ & $87.6\pm0.8$ & $46.3\pm0.7$ \\
TCP \cite{yao2024tcp} & $66.8\pm0.6$ & $39.1\pm0.4$ & $82.6\pm0.4$ & $36.6\pm0.9$ \\
AAPL \cite{kim2024aapl} & $67.4\pm0.8$ & $43.0\pm2.2$ & $84.7\pm0.1$ & $39.8\pm2.0$ \\
PLOT++ \cite{chen2023plot} & $69.5\pm0.5$ & $45.3\pm1.9$ & $86.6\pm0.5$ & $44.3\pm1.0$ \\
{DCPL}  \cite{cao2024domain}& $70.4\pm1.8$ & $49.6\pm1.2$ & $87.9\pm1.7$ & $48.6\pm2.7$ \\ ViP \cite{fang2024aligning} & $69.1\pm1.5$ & $43.0\pm2.8$ & $83.9\pm1.8$ & $43.5\pm4.1$ \\  XCoOp \cite{bie2024xcoop} & $ 71.4 \pm 0.2$ &  $ 53.4 \pm 1.5$ &  $ 89.0 \pm 0.0$ &  $ 49.1 \pm 0.8$
 \\ DePT \cite{zhang2024dept} & \second{$73.0\pm0.4$} & \second{$50.7\pm1.1$} & \second{$90.2\pm0.3$} & \second{$52.3\pm1.0$} \\
CoPrompt \cite{roy2024consistency} & {$71.9\pm2.0$} & {$49.8\pm4.1$} & {$89.0\pm0.4$} & {$51.8\pm4.1$} \\ 
\midrule
\textbf{CILMP (Ours)} & \best{$76.4\pm1.8$} & \best{$58.3\pm0.8$} & \best{$91.9\pm0.5$} & \best{$59.6\pm3.5$} \\
\bottomrule
\end{tabular}
 \caption{Derm7pt}
\end{subtable}}
\resizebox{0.49\textwidth}{!}{
\begin{subtable}{.53\linewidth} \centering
\begin{tabular}{l|cccc}
\toprule
\textbf{Method}           & \textbf{ACC} [\%]   & \textbf{F1} [\%]    & \textbf{AUC} [\%]   & \textbf{Kappa} [\%] \\ 
\midrule
CoOp \cite{zhou2022learning} & $81.4\pm0.9$ & $76.9\pm1.0$ & $96.9\pm0.3$ & $76.0\pm1.1$ \\
CoCoOp \cite{zhou2022conditional} & $85.4\pm0.4$ & $82.3\pm0.8$ & $97.8\pm0.1$ & $81.4\pm0.5$ \\
KgCoOp \cite{yao2023visual} & $64.0\pm0.2$ & $49.4\pm0.5$ & $89.3\pm0.2$ & $52.4\pm0.4$ \\
CLIP-Adapter \cite{gao2024clip} & $78.2\pm0.2$ & $66.3\pm0.4$ & $95.3\pm0.1$ & $71.7\pm0.2$ \\
RPO \cite{lee2023read} & $88.1\pm0.8$ & $86.1\pm0.8$ & $98.5\pm0.1$ & $84.8\pm1.0$ \\
LASP \cite{bulat2023lasp} & $88.4\pm0.3$ & $86.0\pm0.3$ & $98.6\pm0.1$ & $85.2\pm0.3$ \\
PromptSRC \cite{khattak2023self} & $91.4\pm0.4$ & {$90.4\pm0.3$} & {$99.2\pm0.1$} & $89.2\pm0.5$ \\
TCP \cite{yao2024tcp} & $66.8\pm0.4$ & $53.0\pm0.4$ & $91.5\pm0.1$ & $56.3\pm0.5$ \\
AAPL \cite{kim2024aapl} & $83.9\pm0.7$ & $79.5\pm0.8$ & $97.4\pm0.2$ & $79.3\pm0.9$ \\
PLOT++ \cite{chen2023plot} & $86.6\pm0.3$ & $82.4\pm0.4$ & $98.3\pm0.0$ & $82.8\pm0.3$ \\
{DCPL}  \cite{cao2024domain}& $91.9\pm0.3$ & \second{$91.9\pm0.4$} & \second{$99.3\pm0.1$} & \second{$90.8\pm0.4$} \\ ViP \cite{fang2024aligning}  & $83.8\pm0.2$ & $80.5\pm0.6$ & $97.3\pm0.0$ & $79.2\pm0.3$\\  XCoOp \cite{bie2024xcoop} & $ 90.3 \pm 0.6$ &  $ 89.0 \pm 0.6$ &  $ 98.9 \pm 0.0$ &  $ 87.7 \pm 0.7$
 \\ DePT \cite{zhang2024dept} & \second{$92.0\pm0.5$} & {$90.7\pm0.9$} & {$99.2\pm0.0$} & {$89.8\pm0.6$} \\
CoPrompt \cite{roy2024consistency} & $88.9\pm1.2$ & $87.4\pm1.4$ & $98.8\pm0.1$ & $85.8\pm1.6$ \\ 
\midrule
\textbf{CILMP (Ours)} & \best{$93.4\pm0.2$} & \best{$92.7\pm0.2$} & \best{$99.4\pm0.0$} & \best{$91.6\pm0.2$} \\
\bottomrule
\end{tabular}
 \caption{Fetal-US}
\end{subtable}}
\vfill
\resizebox{0.49\textwidth}{!}{
\begin{subtable}{.53\linewidth} \centering
\begin{tabular}{l|cccc}
\toprule
\textbf{Method   }           & \textbf{ACC} [\%]   & \textbf{F1} [\%]    & \textbf{AUC} [\%]   & \textbf{Kappa} [\%] \\

\midrule
CoOp \cite{zhou2022learning} & $75.5\pm0.5$ & $53.8\pm1.2$ & $94.0\pm0.3$ & $54.7\pm1.0$ \\
CoCoOp \cite{zhou2022conditional} & $77.8\pm0.2$ & $59.8\pm0.2$ & $95.1\pm0.2$ & $60.2\pm0.3$ \\
KgCoOp \cite{yao2023visual} & $66.9\pm0.3$ & $27.0\pm0.5$ & $83.4\pm0.4$ & $31.2\pm0.9$ \\
CLIP-Adapter \cite{gao2024clip} & $73.3\pm0.3$ & $41.6\pm0.6$ & $91.7\pm0.2$ & $50.3\pm0.6$ \\
RPO \cite{lee2023read} & $80.4\pm0.5$ & $62.4\pm2.4$ & $95.5\pm0.2$ & $65.1\pm0.9$ \\
LASP \cite{bulat2023lasp} & $79.9\pm0.4$ & $62.8\pm1.5$ & $95.4\pm0.2$ & $64.1\pm0.9$ \\
PromptSRC \cite{khattak2023self} & $81.3\pm0.5$ & $63.6\pm2.2$ & $95.8\pm0.1$ & $66.0\pm1.1$ \\
TCP \cite{yao2024tcp} & $69.4\pm0.1$ & $30.5\pm0.4$ & $86.7\pm0.3$ & $38.5\pm0.1$ \\
AAPL \cite{kim2024aapl} & $76.4\pm0.5$ & $56.7\pm0.2$ & $93.5\pm0.7$ & $56.9\pm0.3$ \\
PLOT++ \cite{chen2023plot} & $80.7\pm0.3$ & {$64.8\pm1.3$} & $95.9\pm0.0$ & $65.8\pm0.5$  \\
{DCPL}  \cite{cao2024domain}& \second{$85.0\pm0.8$} & \second{$74.2\pm1.7$} & \second{$97.4\pm0.2$} & \second{$74.0\pm1.4$}  \\ ViP \cite{fang2024aligning}& $77.2\pm0.2$ & $58.3\pm0.8$ & $94.5\pm0.1$ & $58.8\pm0.4$  \\ XCoOp \cite{bie2024xcoop} & $ 82.9 \pm 0.2$ &  $ 71.3 \pm 0.8$ &  $ 96.8 \pm 0.0$ &  $ 70.2 \pm 0.3$
 \\ DePT \cite{zhang2024dept} & {$83.6\pm0.5$} & {$65.7\pm1.2$} & $95.9\pm0.3$ & {$70.6\pm0.2$} \\
CoPrompt \cite{roy2024consistency} & $79.0\pm0.3$ & $62.0\pm0.7$ & {$96.0\pm0.1$} & $62.3\pm1.0$ \\ 
\midrule
\textbf{CILMP (Ours)} & \best{$86.9\pm0.4$} & \best{$77.4\pm1.0$} & \best{$97.9\pm0.0$} & \best{$77.5\pm0.8$} \\
\bottomrule
\end{tabular}
 \caption{ISIC 2018}
\end{subtable}\hfill}
\resizebox{0.49\textwidth}{!}{
\begin{subtable}{.53\linewidth} \centering
\begin{tabular}{l|cccc}
\toprule
\textbf{Method   }           & \textbf{ACC} [\%]   & \textbf{F1} [\%]    & \textbf{AUC} [\%]   & \textbf{Kappa} [\%] \\ \midrule
CoOp \cite{zhou2022learning} & $84.3\pm0.8$ & $84.2\pm0.8$ & $98.7\pm0.0$ & $82.0\pm0.9$ \\
CoCoOp \cite{zhou2022conditional} & $86.5\pm0.2$ & $86.5\pm0.2$ & $99.0\pm0.0$ & $84.5\pm0.3$ \\
KgCoOp \cite{yao2023visual} & $74.3\pm0.1$ & $74.5\pm0.1$ & $97.0\pm0.1$ & $70.6\pm0.1$ \\
CLIP-Adapter \cite{gao2024clip} & $83.6\pm0.3$ & $83.5\pm0.3$ & $98.6\pm0.0$ & $81.2\pm0.4$ \\
RPO \cite{lee2023read} & $87.3\pm0.4$ & $87.3\pm0.4$ & $99.1\pm0.1$ & $85.5\pm0.5$ \\
LASP \cite{bulat2023lasp} & $89.1\pm0.2$ & $89.1\pm0.2$ & $99.2\pm0.0$ & $87.6\pm0.2$ \\
PromptSRC \cite{khattak2023self} & {$91.6\pm0.2$} & {$91.6\pm0.2$} & $99.4\pm0.1$ & {$90.5\pm0.2$} \\
TCP \cite{yao2024tcp} & $78.5\pm0.5$ & $78.6\pm0.5$ & $97.7\pm0.1$ & $75.4\pm0.6$ \\
AAPL \cite{kim2024aapl} & $85.8\pm0.5$ & $85.8\pm0.5$ & $98.9\pm0.0$ & $83.8\pm0.5$ \\
PLOT++ \cite{chen2023plot} & $89.2\pm0.2$ & $89.2\pm0.2$ & $99.3\pm0.0$ & $87.7\pm0.2$ \\
{DCPL}  \cite{cao2024domain}& \second{$92.4\pm0.3$} & \second{$92.4\pm0.3$} & \best{$99.6\pm0.0$} & \second{$91.4\pm0.3$} \\ ViP \cite{fang2024aligning} & $86.5\pm0.3$ & $86.5\pm0.2$ & $98.9\pm0.0$ & $84.5\pm0.3$\\  XCoOp \cite{bie2024xcoop} & $ 90.9 \pm 0.3$ &  $ 90.9 \pm 0.4$ &  $ 99.5 \pm 0.0$ &  $ 89.6 \pm 0.4$
\\ DePT \cite{zhang2024dept} & {$91.0\pm0.5$} & {$90.9\pm0.5$} & $99.4\pm0.0$ & $89.7\pm0.6$ \\
CoPrompt \cite{roy2024consistency} & {$91.6\pm0.4$} & {$91.6\pm0.4$} & \second{$99.5\pm0.0$} & $90.4\pm0.5$ \\ 
\midrule
\textbf{CILMP (Ours)} & \best{$93.2\pm0.1$} & \best{$93.2\pm0.1$} & \best{$99.6\pm0.0$} & \best{$92.2\pm0.1$} \\
\bottomrule
\end{tabular}
 \caption{Kvasir}
\end{subtable}}
\vfill
\resizebox{0.49\textwidth}{!}{
\begin{subtable}{.53\linewidth} \centering
\begin{tabular}{l|cccc}
\toprule
\textbf{Method   }           & \textbf{ACC} [\%]   & \textbf{F1} [\%]    & \textbf{AUC} [\%]   & \textbf{Kappa} [\%] \\ \midrule
CoOp \cite{zhou2022learning} & $56.6\pm0.3$ & $37.2\pm1.3$ & $81.3\pm0.5$ & $29.0\pm0.4$ \\
CoCoOp \cite{zhou2022conditional} & $56.0\pm0.6$ & $37.3\pm2.4$ & $80.6\pm0.7$ & $27.8\pm1.9$ \\
KgCoOp \cite{yao2023visual} & $49.6\pm0.3$ & $21.2\pm1.0$ & $69.5\pm0.1$ & $11.8\pm0.3$ \\
CLIP-Adapter \cite{gao2024clip} & $49.9\pm0.4$ & $23.5\pm1.1$ & $73.2\pm0.7$ & $13.5\pm0.9$ \\
RPO \cite{lee2023read} & $59.2\pm0.6$ & $44.9\pm0.5$ & $83.5\pm0.3$ & $35.8\pm0.3$ \\
LASP \cite{bulat2023lasp} & $60.6\pm0.5$ & $45.6\pm1.2$ & $83.6\pm0.7$ & $38.9\pm1.2$ \\
PromptSRC \cite{khattak2023self} & $60.3\pm1.9$ & $41.5\pm4.2$ & $83.9\pm0.5$ & $36.3\pm3.7$ \\
TCP \cite{yao2024tcp} & $51.4\pm0.3$ & $25.7\pm0.7$ & $73.7\pm0.3$ & $16.2\pm0.6$ \\
AAPL \cite{kim2024aapl} & $56.1\pm0.6$ & $37.1\pm0.5$ & $80.3\pm0.1$ & $28.5\pm1.0$ \\
PLOT++ \cite{chen2023plot} & $58.7\pm0.6$ & $44.9\pm0.9$ & $84.2\pm0.2$ & $37.3\pm0.9$ \\
{DCPL}  \cite{cao2024domain} & $66.6\pm0.8$ & \second{$55.7\pm1.3$} & \second{$89.7\pm0.4$} & $49.5\pm1.2$ \\ ViP \cite{fang2024aligning} & $57.4\pm0.1$ & $38.3\pm0.8$ & $81.3\pm0.5$ & $30.8\pm0.5$\\  XCoOp \cite{bie2024xcoop} & $ 62.9 \pm 0.5$ &  $ 50.0 \pm 0.3$ &  $ 87.2 \pm 0.2$ &  $ 42.7 \pm 0.6$
\\ DePT \cite{zhang2024dept} & $62.5\pm0.5$ & $44.2\pm3.2$ & $85.2\pm0.7$ & $42.2\pm1.8$ \\
CoPrompt \cite{roy2024consistency} & \second{$67.6\pm0.7$} & {$55.6\pm0.3$} & \second{$89.7\pm0.6$} & \second{$51.5\pm0.8$} \\ 
\midrule
\textbf{CILMP (Ours)} & \best{$68.5\pm0.9$} & \best{$57.9\pm1.3$} & \best{$90.7\pm0.3$} & \best{$53.4\pm1.4$} \\
\bottomrule
\end{tabular}
 \caption{ODIR}
\end{subtable}\hfill}
\resizebox{0.49\textwidth}{!}{
\begin{subtable}{.53\linewidth} \centering
\begin{tabular}{l|cccc}
\toprule
\textbf{Method   }           & \textbf{ACC} [\%]   & \textbf{F1} [\%]    & \textbf{AUC} [\%]   & \textbf{Kappa} [\%] \\ \midrule
CoOp \cite{zhou2022learning} & $75.9\pm0.6$ & $74.6\pm1.0$ & $90.4\pm0.2$ & $62.6\pm0.9$ \\
CoCoOp \cite{zhou2022conditional} & $77.8\pm1.3$ & $76.3\pm2.0$ & $91.4\pm0.6$ & $65.6\pm2.2$ \\
KgCoOp \cite{yao2023visual} & $63.6\pm0.8$ & $61.7\pm0.6$ & $79.0\pm0.5$ & $42.9\pm1.2$ \\
CLIP-Adapter \cite{gao2024clip} & $72.1\pm0.5$ & $69.3\pm0.8$ & $88.3\pm0.1$ & $56.2\pm0.9$ \\
RPO \cite{lee2023read} & $85.1\pm0.5$ & $83.8\pm0.7$ & $95.0\pm0.3$ & $77.2\pm0.8$ \\
LASP \cite{bulat2023lasp} & $84.5\pm0.3$ & $83.3\pm0.2$ & $94.5\pm0.1$ & $76.2\pm0.4$ \\
PromptSRC \cite{khattak2023self} & $84.5\pm1.0$ & $83.6\pm1.0$ & $95.2\pm0.4$ & $76.6\pm1.4$ \\
TCP \cite{yao2024tcp} & $67.2\pm0.1$ & $65.2\pm0.4$ & $82.4\pm0.3$ & $48.5\pm0.3$ \\
AAPL \cite{kim2024aapl} & $77.2\pm1.4$ & $75.6\pm1.6$ & $91.1\pm0.5$ & $64.7\pm2.3$ \\
PLOT++ \cite{chen2023plot} & {$86.9\pm0.4$} & {$85.6\pm0.4$} & {$95.8\pm0.1$} & {$79.8\pm0.5$} \\
{DCPL}  \cite{cao2024domain} & \second{$87.5\pm0.2$} & \second{$86.1\pm0.3$} & \second{$96.0\pm0.3$} & \second{$80.8\pm0.4$}
\\  ViP \cite{fang2024aligning}& $79.0\pm0.5$ & $78.6\pm0.5$ & $89.7\pm0.5$ & $68.0\pm0.7$ \\  XCoOp \cite{bie2024xcoop} & $ 84.4 \pm 0.7$ &  $ 83.2 \pm 0.8$ &  $ 94.9 \pm 0.3$ &  $ 76.0 \pm 1.0$
\\ DePT \cite{zhang2024dept} & $85.0\pm0.5$ & $84.1\pm0.5$ & $95.5\pm0.7$ & $77.4\pm0.8$ \\
CoPrompt \cite{roy2024consistency} & $80.1\pm0.3$ & $79.1\pm0.5$ & $93.2\pm0.5$ & $70.0\pm0.4$ \\ 
\midrule
\textbf{CILMP (Ours)} & \best{$89.5\pm0.7$} & \best{$88.7\pm0.7$} & \best{$96.7\pm0.3$} & \best{$84.0\pm1.0$} \\
\bottomrule
\end{tabular}
 \caption{Pneumonia}
\end{subtable}}
\vspace{-2.5em}
\end{table*}

\subsection{Implementation Details}
For a fair comparison, all experiments are conducted using the pre-trained CLIP model \cite{radford2021learning}. Specifically, the image encoder employs the ViT-B/16 backbone \cite{dosovitskiy2020image}, which has an output dimensionality of 512. The text encoder utilizes the Transformer backbone, incorporating architectural refinements as detailed in \cite{radford2019language}.
In our CILMP framework, we utilize the LLaMA-3-8B \cite{llama3modelcard} as the LLM, which is one of the most capable openly available LLM to date.
LLaMA-3-8B is a decoder-only LLM, which has 32 layers with a representation dimension of 4096.
The length of contexts for image prompts and text prompts is set to 4.
In alignment with existing methods \cite{khattak2023self, zhang2024dept, chen2023plot, roy2024consistency}, we employ a deep prompting strategy for both encoders, injecting the prompts into intermediate layers. 
Prompts are initialized with a zero-mean Gaussian distribution with a standard deviation of 0.02, except for the text prompts in the first layer, which are initialized with the word embeddings of ``a photo of a".
The model is trained for 100 epochs with a batch size of 64. 
During training, an SGD optimizer is employed with an initial learning rate of 0.0025.
Training hyper-parameters are kept consistent across datasets unless otherwise specified.
To ensure fair comparisons, we train all the compared methods with the same configurations, including the training epochs, batch size, and model backbones.
All models are trained using one NVIDIA RTX 4090 GPU {(24GB)}.
We report the performance of the model from the last epoch, and no validation set is used during the training process.
%
%
For the image inputs, although data augmentation is very useful to improve performance in medical image analysis, we simply follow \cite{zhou2022learning, zhou2022conditional, zhang2024dept, khattak2023self} to apply random resized crop and random flip as the data augmentations to maintain fair comparisons.
The input size of training images is $224\times 224$.
During the inference stage, the size of test images is also set to $224 \times 224$.
For the evaluation of the methods, we employ four widely recognized metrics, including the Accuracy, F1-score, area under the ROC curve (AUROC), and Kappa score. 
These metrics are chosen to provide a comprehensive assessment of the model performance across different aspects.
The results are reported as the mean and standard deviation over three independent runs with different random seeds.

\begin{table}[t]
\centering
\caption{{Comparison with medical vision language foundation models. ``ZS" means zero-shot performance, ``FFT" means fully fine-tuning paradigm, which fine-tunes all parameters of the VLM. ``PT" denotes the parameter-efficient fine-tuning. ``Param." is the number of trainable parameters.}}
\label{tab:vlm}
\renewcommand\arraystretch{1.0}
\resizebox{0.49\textwidth}{!}{
{
\begin{tabular}{c|l|c|cccc} \toprule
\textbf{Setup} & \textbf{Method}                & \textbf{Param.} & \textbf{ACC} {[}\%{]} & \textbf{F1} {[}\%{]} & \textbf{AUC} {[}\%{]} & \textbf{Kappa} {[}\%{]} \\ \midrule
\multicolumn{7}{l}{{{\textbf{\textit{Dermatoscopy}:}}}}
 \\ \midrule
\multirow{3}{*}{\rotatebox{90}{ZS}} & PubMedCLIP  \cite{eslami2023pubmedclip}        & -    &  10.1         &  6.4            &     57.7        &    2.3                          \\
& BioMedCLIP  \cite{zhang2023biomedclip}  &  -         & 43.8          &      16.7         & 67.8               & 10.3 \\
& MONET  \cite{kim2024transparent}  &    -       &     54.2         &    29.4         &  80.9            & 28.3 \\ \midrule
\multirow{3}{*}{\rotatebox{90}{FFT}}  & PubMedCLIP   \cite{eslami2023pubmedclip}           &  151M          &   85.7$\pm$0.3            &    55.4$\pm$0.5          &     94.8$\pm$0.1          &     72.3$\pm$0.9            \\
& BioMedCLIP \cite{zhang2023biomedclip}   &  196M         &  87.5$\pm$0.2            &    80.1$\pm$1.1          &     98.0$\pm$0.1          &     76.0$\pm$0.5  \\

 & MONET \cite{kim2024transparent}  &  428M         &  88.3$\pm$1.1            &     79.3$\pm$1.8        &  98.1$\pm$0.2            & 77.6$\pm$1.8 \\ \midrule
\multirow{1}{*}{\rotatebox{90}{PT}} & \textbf{CILMP (Ours)}    &  3.8M         &     87.4$\pm$0.3         & 77.7$\pm$1.0            & 98.0$\pm$0.0             &       76.0$\pm$0.4        \\ 
 \midrule
\multicolumn{7}{l}{{{\textbf{\textit{Chest X-ray}:}}}}
 \\ \midrule
\multirow{3}{*}{\rotatebox{90}{ZS}} & PubMedCLIP    \cite{eslami2023pubmedclip}          & -          &  39.4            &    27.3         &  61.7            & 8.4               \\
& BioMedCLIP  \cite{zhang2023biomedclip} &       -    &    51.6          &   45.6          &  75.6            & 22.2\\
& MONET  \cite{kim2024transparent}  &    -       &    51.8          &   39.4          &  59.8            & 22.7 \\ \midrule
\multirow{3}{*}{\rotatebox{90}{FFT}} & PubMedCLIP  \cite{eslami2023pubmedclip}           &    151M        &   90.5$\pm$0.7           &  89.7$\pm$0.7           & 97.1$\pm$0.2             &   85.5$\pm$1.0             \\

& BioMedCLIP \cite{zhang2023biomedclip}  & 196M       &    90.4$\pm$0.1           &  89.4$\pm$0.1           & 97.2$\pm$0.1             &   85.2$\pm$0.2  \\
& MONET \cite{kim2024transparent}  &  428M         &   88.8$\pm$0.5           &       87.8$\pm$0.6      & 97.0$\pm$0.3             &82.8$\pm$0.7  \\ \midrule
\multirow{1}{*}{\rotatebox{90}{PT}} & \textbf{CILMP (Ours)}   & 3.8M  & {89.5$\pm$0.7} & {88.7$\pm$0.7} & {96.7$\pm$0.3} & {84.0$\pm$1.0}           \\ 
\bottomrule
\end{tabular}}}
\vspace{-1em}
\end{table}

\subsection{Comparison with the State-of-the-art (SOTA) Methods}
In TABLEs~\ref{tab-sota-results1} and \ref{tab-sota-results2}, we compare our method with {15} SOTA prompt tuning methods over 11 datasets.
The compared benchmarking methods include CoOp~\cite{zhou2022learning}, CoCoOp~ \cite{zhou2022conditional}, KgCoOp~\cite{zhang2023knowledge}, CLIP-Adapter~\cite{gao2024clip}, PRO~\cite{lee2023read}, LASP~\cite{bulat2023lasp}, PromptSRC~\cite{khattak2023self}, TCP~\cite{yao2024tcp}, AAPL~\cite{kim2024aapl}, PLOT++~\cite{chen2023plot}, {DCPL}~\cite{cao2024domain}, {ViP}~\cite{fang2024aligning}, {XCoOp}~\cite{bie2024xcoop},
DePT~\cite{zhang2024dept} and CoPrompt~\cite{roy2024consistency}.
{DCPL introduces MedSAM~\cite{ma2024segment} as a knowledge provider, KgCoOp introduces hand-crafted knowledge into prompt tuning, and ViP, XCoOp and CoPrompt employ LLM to provide fixed knowledge.}
{It is noteworthy that most baseline prompt tuning methods (with the exception of ViP and XCoOp) are specifically designed for few-shot image classification, whereas our work focuses on the full-shot training scenario.}

Our method significantly outperforms milestone prompt tuning methods such as CoOp, CoCoOp, and CLIP-Adapter across 11 datasets. 
In the fundus modality, our method surpasses all recently proposed advanced prompt tuning methods, such as PromptSRC, DePT, and CoPrompt. 
Notably, on the challenging ODIR dataset, our method outperforms the CoPrompt method, which already exhibits high performance, by 0.9\% in accuracy, 2.3\% in F1 score, 1.0\% in AUC score, and 1.9\% in kappa score. 
A similar trend is observed in the dermatoscope modality with the DermaMNIST, Derm7pt, and ISIC datasets. 
Across these three datasets, our method exceeds the DePT method, which is built on top of the soild PromptSRC method, by 3.5\% in accuracy and 2.0\% in AUC score on average. 
Moreover, in the ultrasound modality on the Fetal-US dataset, our method shows a performance improvement of 1.4\%, 2.0\%, 0.2\%, and 1.8\% in each metric over the second-best DePT method. 
Furthermore, on the category-balanced Kvasir dataset of the endoscope modality, our method outperforms the CoPrompt method by 1.6\% in accuracy, 1.6\% in F1-score, 0.1\% in AUC score, and 1.8\% in kappa score.
X-ray images present a particular challenge for our method. 
{
While it surpasses other methods on the Pneumonia dataset by over 2.0\% in accuracy, 2.6\% in F1-score, 0.7\% in AUC score, and 3.2\% in kappa score, it lags behind on the CPN-X-ray dataset. }
On CPN-X-ray, the CoPrompt method achieves the best performance with an accuracy of 97.7\%, an F1-score of 97.8\%, an AUC of 99.8\%, and a kappa score of 96.5\%. 
In comparison, our method underperforms CoPrompt by 1.5\% in accuracy, 1.5\% in F1-score, 0.2\% in AUC score, and 1.2\% in kappa score. 
This underperformance is attributed to CoPrompt's use of more data augmentations and the combination of different PEFT strategies, including both prompt tuning and adapter, which provide increased tuning flexibility.
Nevertheless, these results highlight the competitiveness of our proposed method compared to recent prompt tuning methods.

{We also compare CILMP with three prompt tuning methods specifically designed for the medical domain, including DCPL, ViP and XCoOp.}
{Notably, for the DCPL method, MedSAM}~\cite{ma2024segment} {is employed as the domain-specific feature encoder to align with the original paper. }
{As shown in} \ref{tab-sota-results1} and \ref{tab-sota-results2}, {our CILMP outperforms these methods by significant margins across almost all datasets.
For instance, on the ISIC dataset, CILMP exceeds DCPL, the second-best performing method in this modality, by 1.9\%, 3.2\%, 0.5\%, and 3.5\% in terms of accuracy, F1 score, AUC score, and Kappa score, respectively,
Furthermore, in the fundus modality, CILMP surpasses XCoOp by 5.6\%, 7.9\%, 3.5\%, and 10.7\% for each metric.
These comparisons further highlight the effectiveness of the CILMP method in the field of medical vision-language model research.}

{In summary}, TABLE \ref{tab:average} presents a more comprehensive comparison, which includes the average results in 11 datasets.
{As illustrated, our CILMP method consistently outperforms the second-best approach DCPL by 1.7\%, 2.4\%, 0.9\%, and 3.1\% in terms of accuracy, F1-score, AUC score, and Kappa score, respectively. 
Comparison with other knowledge-enhanced methods, such as KgCoOp, XCoOp, and CoPrompt, also shows our method's superiority in the medical image classification tasks.}
Overall, these results indicate that our method demonstrates superior performance across diverse data modalities and distributions, highlighting its enhanced capability to transfer pre-trained vision language models for medical image classification over existing state-of-the-art prompt tuning methods.
Consequently, the experimental results affirm the effectiveness of the proposed method.

\subsection{{Comparison with Medical Vision Language Models}}

TABLE~\ref{tab:vlm} {compares our method with more medical vision language models.
Notably, the focus of our work is to design effective adaptation techniques to adapt VLM for downstream datasets. 
In contrast, existing medical VLMs aim to achieve strong transfer capabilities through extensive pre-training on large-scale datasets, {which is an orthogonal line to our study.}
The experiments show that even state-of-the-art VLMs pre-trained on massive medical datasets exhibit limited zero-shot performance on dermatoscopy and chest X-ray tasks (\textit{e.g.},} MONET~\cite{kim2024transparent} {achieves only 54.2\% ACC and 80.9\% AUC on dermatoscopy despite in-domain pretraining), underscoring the necessity of task-specific adaptation in clinical applications.
{Then}, when fully fine-tuned, these VLMs can achieve strong performance (\textit{e.g.}, 88.3\% ACC for MONET on dermatoscopy), but at the cost of updating a large number of parameters (\textit{e.g.}, 428M for MONET).
In contrast, our method with merely 3.8M trainable parameters delivers competitive results.
For example, on chest X-ray, CILMP attains 89.5\% ACC, trailing fully fine-tuned} PubMedCLIP~\cite{eslami2023pubmedclip} and BioMedCLIP~\cite{zhang2023biomedclip} {by only about 1\% while reducing the trainable parameters by 40–50x.
Overall, CILMP not only outperforms zero-shot medical VLMs by large margins but also narrows the gap to the fully fine-tuning paradigm with extreme parameter efficiency, demonstrating its effectiveness in bridging general-purpose medical VLMs to domain-specific clinical tasks.
}

\begin{table}[t!]
\centering
\caption{Ablation study on the effectiveness of each component in CILMP {on the dermatoscopy modality.}
``RD" means the relationship descriptor. 
}
\label{tab-main-ablation}
\setlength\tabcolsep{4.5pt}
\renewcommand\arraystretch{1.0}
\resizebox{0.49\textwidth}{!}{
\begin{tabular}{c|l|cccc} \toprule
\textbf{Dataset}                     & \textbf{Strategy}           & \textbf{ACC} [\%] & \textbf{F1} [\%] & \textbf{AUC} [\%] & \textbf{Kappa} [\%] \\ \midrule
\multirow{4}{*}{\rotatebox{90}{DermaMN}} & CILMP              & 87.43$\pm$0.33        & 77.73$\pm$0.97       & 98.03$\pm$0.05        & 76.03$\pm$0.45          \\
                            & \ \textit{w/o} RD             & 87.00$\pm$0.16        & 76.70$\pm$0.93       & 97.77$\pm$0.12        & 75.10$\pm$0.29         \\
                            & \ \textit{w/o}   Conditional  & 86.63$\pm$0.29        & 76.30$\pm$1.27       & 97.80$\pm$0.16        & 74.70$\pm$0.57          \\
                            & \ \textit{w/o}   Intervention & 86.00$\pm$0.49        & 74.57$\pm$0.17       & 97.40$\pm$0.22        & 72.93$\pm$1.07          \\
                            \midrule
\multirow{4}{*}{\rotatebox{90}{Derm7pt}}
& CILMP              & 76.40$\pm$1.76        & 58.27$\pm$0.76       & 91.93$\pm$0.49        & 59.63$\pm$3.48          \\
                            & \ \textit{w/o} RD             & 75.43$\pm$0.56        & 56.83$\pm$1.11       & 91.40$\pm$0.29        & 58.67$\pm$0.82          \\
                            & \ \textit{w/o} Conditional    & 73.00$\pm$0.85        & 51.83$\pm$1.32       & 90.93$\pm$0.09        & 53.03$\pm$2.15          \\
                            & \ \textit{w/o} Intervention   & 70.13$\pm$1.43        & 47.23$\pm$1.96       & 87.03$\pm$0.39        & 47.83$\pm$3.53          \\
                            \midrule
\multirow{4}{*}{\rotatebox{90}{ISIC 2018}}
& CILMP              & 86.90$\pm$0.42        & 77.40$\pm$0.99       & 97.90$\pm$0.00        & 77.47$\pm$0.75          \\
                            & \ \textit{w/o} RD             & 86.27$\pm$0.75        & 75.80$\pm$1.13       & 97.83$\pm$0.09        & 76.00$\pm$1.35          \\ 
                            & \ \textit{w/o} Conditional    & 85.67$\pm$0.21        & 75.57$\pm$0.78       & 97.47$\pm$0.09        & 75.03$\pm$0.25          \\
                            & \ \textit{w/o} Intervention   & 85.33$\pm$0.66        & 74.93$\pm$1.51       & 97.47$\pm$0.09        & 74.33$\pm$1.23          \\
                          \midrule
\multirow{4}{*}{\rotatebox{90}{Average}}& CILMP              & 83.58        & 71.13       & 95.95        & 71.04          \\
                            & \ \textit{w/o} RD             & 82.90        & 69.78       & 95.67        & 69.92          \\ 
                            & \ \textit{w/o} Conditional    & 81.77        & 67.90       & 95.40        & 67.59          \\
                            & \ \textit{w/o} Intervention   & 80.49        & 65.58       & 93.97        & 65.03          \\
                           \bottomrule
\end{tabular}}
\vspace{-0.5em}
\end{table}

\subsection{Ablation Study}
\subsubsection{Ablation study of CILMP components}
We investigate the effectiveness of three key components in CILMP: the relationship descriptor (\textit{w/o} RD), the conditional mechanism (\textit{w/o} Conditional), and the intervention function (\textit{w/o} Intervention).
The experimental results are presented in TABLE~\ref{tab-main-ablation}. Experiments are conducted on the dermatoscope image modality using three datasets: DermaMNIST, Derm7pt, and ISIC 2018, chosen for their representativeness and challenging classification tasks due to multiple categories and varying training set scales.
Firstly, removing the relationship descriptor (\textit{w/o} RD in TABLE~\ref{tab-main-ablation}) means the model merely conditions the intervention on the input image without injecting the matching prior. This leads to notable performance drops across each dataset. On average, the accuracy, F1 score, AUC, and Kappa decrease by 0.70\%, 1.32\%, 0.28\%, and 1.12\%, respectively.
Removing the conditional mechanism (\textit{w/o} Conditional in TABLE~\ref{tab-main-ablation}) reverts to using the original intervention function (Eq.~(\ref{eq:11})) proposed in \cite{wu2024reft}.
The exclusion of the conditional mechanism results in accuracy decreases of 0.80\%, 3.40\%, and 1.23\% across the datasets. Similar declines are observed in the F1 and Kappa scores, each dropping by over 2\%. Although the AUC score remains relatively stable, likely due to its already high value. 
Moreover, removing the intervention function $\Psi$ (\textit{w/o} Intervention) reduces the framework to merely projecting LLM representations into the prompt dimension and concatenating them to form class-specific prompts for the VLM. 
The results show that this degradation underperforms our CILMP method by an average of 3.09\%, 5.55\%, 1.98\%, and 5.01\% in each metric across three datasets, respectively.
{To further demonstrate the generalizability of the proposed components across diverse medical imaging domains, we then conduct ablation study on the fundus modality, which comprises 3 datasets including ADAM, APTOS and ODIR.
The results are shown in TABLE} \ref{tab-main-ablation-fundus}.
{As can be seen,  the omission of the
conditional mechanism results in a performance decrease of 3.28\%, whereas the absence of the proposed intervention function leads to an average performance decline of 3.96\% across the three datasets in terms of F1 score.}
In summary, these findings underscore the critical importance of the intervention function $\Psi$, the conditional mechanism, and the relationship descriptor in adapting LLM representations to the VLM space and generating class-specific, instance-adaptive prompts, thereby significantly enhancing the performance of the CILMP framework.

\begin{table}[t!]
\centering
\caption{{Ablation study on the effectiveness of each component in CILMP on the fundus modality. }
}
\label{tab-main-ablation-fundus}
\setlength\tabcolsep{4.5pt}
\renewcommand\arraystretch{0.95}
\resizebox{0.49\textwidth}{!}{
{
\begin{tabular}{c|l|cccc} \toprule
\textbf{Dataset}                     & \textbf{Strategy}           & \textbf{ACC} [\%] & \textbf{F1} [\%] & \textbf{AUC} [\%] & \textbf{Kappa} [\%] \\ \midrule
\multirow{4}{*}{\rotatebox{90}{ADAM}} & CILMP    &        90.17$\pm$0.47 & 85.47$\pm$0.62 & 93.53$\pm$0.34 & 70.97$\pm$1.21          \\
                            & \ \textit{w/o} RD             & 89.67$\pm$0.47        & 85.17$\pm$0.47       & 94.13$\pm$0.75        & 70.40$\pm$0.99         \\
                            & \ \textit{w/o}   Conditional  & 89.23$\pm$0.29        & 85.03$\pm$0.25       & 93.50$\pm$0.14        & 70.07$\pm$0.50          \\
                            & \ \textit{w/o}   Intervention & 88.15$\pm$1.35        & 83.07$\pm$1.58       & 92.30$\pm$0.21        & 66.15$\pm$3.15          \\
                            \midrule
\multirow{4}{*}{\rotatebox{90}{APTOS}}
& CILMP              & 84.47$\pm$0.46        & 66.67$\pm$0.76       & 94.63$\pm$0.21        & 75.73$\pm$0.71          \\
                            & \ \textit{w/o} RD             & 83.87$\pm$0.61        & 65.67$\pm$1.52       & 94.43$\pm$0.05        & 74.83$\pm$1.11          \\
                            & \ \textit{w/o} Conditional    & 83.27$\pm$0.37        & 62.53$\pm$0.42       & 94.63$\pm$0.05        & 73.90$\pm$0.57          \\
                            & \ \textit{w/o} Intervention   & 82.17$\pm$0.62        & 59.73$\pm$1.48       & 94.17$\pm$0.12        & 71.90$\pm$0.99          \\
                            \midrule
\multirow{4}{*}{\rotatebox{90}{ODIR}}
& CILMP              & 68.47$\pm$0.87        & 57.93$\pm$1.32       & 90.67$\pm$0.31        & 53.43$\pm$1.41          \\
                            & \ \textit{w/o} RD             & 67.80$\pm$1.30        & 57.45$\pm$1.15       & 89.70$\pm$0.30        & 52.65$\pm$2.05         \\ 
                            & \ \textit{w/o} Conditional    & 67.37$\pm$0.57        & 56.67$\pm$0.76       & 89.83$\pm$0.29        & 51.27$\pm$0.58          \\
                            & \ \textit{w/o} Intervention   & 65.97$\pm$1.71        & 55.37$\pm$2.98       & 89.23$\pm$1.16        & 48.63$\pm$3.50          \\
                          \midrule
\multirow{4}{*}{\rotatebox{90}{Average}}& CILMP              & 81.04        & 70.02       & 93.61        & 66.71          \\
                            & \ \textit{w/o} RD             & 80.45        & 69.43       & 92.75        & 65.96          \\ 
                            & \ \textit{w/o} Conditional    & 79.96        & 66.74       & 92.65        & 65.08          \\
                            & \ \textit{w/o} Intervention   & 78.76        & 66.06       & 91.90        & 62.23          \\
                           \bottomrule
\end{tabular}}}
\vspace{-1.5em}
\end{table}

\subsubsection{Ablation on the usage of LLM} 

Different from our CILMP method using LLM representations to incorporate medical domain knowledge during prompt tuning, previous studies \cite{yao2023visual, roy2024consistency} commonly utilize plain text descriptions.
{In this plain text setting, the LLM generates a text description for each disease, which is then combined
with the disease name to create a prompt for the VLM.}
To further explore the effectiveness of these two strategies, we conduct comparable experiments and report the results in TABLE~\ref{tab:LLM-representation}.
On average across three datasets, our representation-based method outperforms text format strategy by 1.80\% in accuracy and 4.24\% in F1 score.
These comparisons demonstrate that using LLM representations in conjunction with our proposed intervention mechanism is a more effective way to incorporate medical domain knowledge into the prompt-tuning process.

\subsubsection{Ablation of the prefix/suffix length of intervention}
We conducted an ablation study to examine the sensitivity of our method to the prefix and suffix lengths (from 2 to 16) for LLM representations, setting $L_{\text{prefix}}=L_{\text{suffix}}$ for simplicity.
As shown in TABLE~\ref{tab:prefix}, optimal performance is achieved with an intervention length of 4. 
Consequently, we set $L_{\text{prefix}} = L_{\text{suffix}} = 4$ across all datasets.
However, it is important to note that this may not be optimal for every dataset and could be further tuned to improve performance.

\subsubsection{Ablation of the dimensionality of the linear subspace}
We conducted experiments on various dimensionalities of the linear subspace (1, 4, 8, and 16 dimensions) and found that a dimensionality of 8 achieves the best performance, as shown in TABLE~\ref{tab:linear-subspace}. Consequently, we set the linear subspace dimensionality to 8 for all other settings.

\subsubsection{{Ablation of the prompt length}}
{We also conducted experiments on the prompt length (1, 4, 8, and 16 context tokens).
As shown in TABLE}~\ref{tab:prompt-length}{, a length of 4 yields optimal performance with low sensitivity to parameter changes.
Consequently, we set the prompt length to 4 for all other settings.}

\begin{table}[t!]
\caption{Ablation study on the usage of large language model. 
``DermaMN" means the DermaMNIST dataset. ``Average" means the average results across three datasets of the dermatoscope modality.}
\label{tab:LLM-representation}
\centering
\setlength\tabcolsep{4.5pt}
\renewcommand\arraystretch{0.95}
\resizebox{0.49\textwidth}{!}{
\begin{tabular}{l|l|cccc} \toprule
\textbf{Dataset}                     & \textbf{Strategy} & \textbf{ACC} [\%]  & \textbf{F1} [\%]   & \textbf{AUC} [\%] & \textbf{Kappa} [\%] \\ \midrule
\multirow{2}{*}{DermaMN} & Text    & 86.33$\pm$0.68 & 74.47$\pm$1.01 & 97.67$\pm$0.17  & 73.87$\pm$1.31  \\
                            & Representation   & 87.43$\pm$0.33 & 77.73$\pm$0.97 & 98.03$\pm$0.05  & 76.03$\pm$0.45    \\ \midrule
\multirow{2}{*}{Derm7pt}    & Text    & 73.83$\pm$0.80 & 51.93$\pm$1.27 & 91.33$\pm$0.83  & 54.60$\pm$1.90  \\
                            & Representation   & 76.40$\pm$1.76 & 58.27$\pm$0.76 & 91.93$\pm$0.49  & 59.63$\pm$3.48  \\ \midrule
\multirow{2}{*}{ISIC 2018}       &  Text   & 85.17$\pm$0.21 & 74.27$\pm$0.39 & 94.93$\pm$0.88  & 74.10$\pm$0.29  \\ 
                            & Representation   & 86.90$\pm$0.42 & 77.40$\pm$0.99 & 97.90$\pm$0.00  & 77.47$\pm$0.75  \\ \midrule
\multirow{2}{*}{Average}     &  Text   & 81.78 & 66.89 & 95.31  & 67.52  \\
                            & Representation   & 83.58 & 71.13 & 95.95  & 71.04   \\ \bottomrule 
\end{tabular}}
\vspace{-0.5em}
\end{table}

\begin{figure}[t!]
\centering
\includegraphics[width=0.37\textwidth]{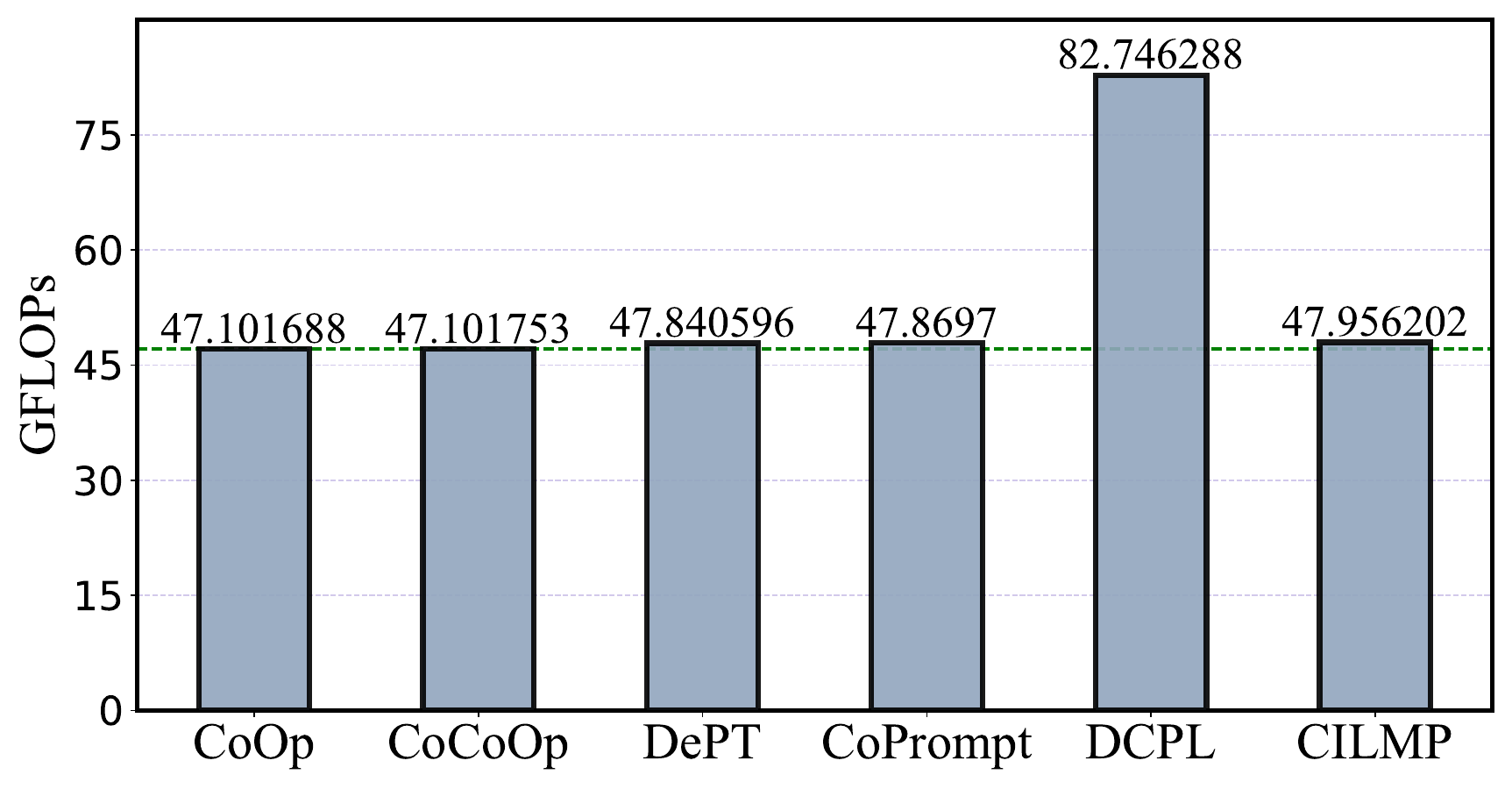}
\vspace{-0.5em}
\caption{{Comparison of our CILMP and the other competitive prompt tuning methods in FLOPs.
Comparison is conducted on the ADAM dataset.}
}
\vspace{-1.5em}
\label{fig:flops}
\end{figure}

\subsection{{Efficiency Analysis}}
\subsubsection{Trainable Parameter Statistics}
{TABLE}~\ref{tab:average} {presents the number of trainable parameters for each prompt tuning method. 
In alignment with established practices in the literature, these methods employ the ViT-B/16 backbone of CLIP.
CoPrompt contributes an additional 4.7M parameters, DCPL adds 3.8M, and ViP incorporates 41.5M parameters. 
In contrast, our proposed CILMP method introduces 3.8M parameters, yet it achieves better performance compared to these approaches.
Moreover, compared to the whole CLIP that consists of 149.6M parameters, our method introduces approximately 2.5\% of trainable parameters relative to the complete VLM. 
Compared to the LLM used, it introduces only about 0.0475\% additional parameters.}

\subsubsection{FLOPs Statistics}
{We also compare the computing efficiency between CILMP and other competitive prompt tuning methods in terms of FLOPs.
The results are shown in Fig.} \ref{fig:flops}.
{Note that CoOp does not introduce computational burden compared to the original CLIP, so it can be viewed as a baseline for other methods.
Compared to CoOp, DCPL incurs a significant increase of 35.6445 GFLOPs, primarily because it requires forwarding MedSAM during both training and inference.
In contrast, our CILMP only adds 0.8545 GFLOPs, which is on par with other methods like CoCoOp, DePT, and CoPrompt, but it delivers significantly better performance.
Although CILMP involves jointly fine-tuning an LLM and a VLM, it does not require any forward or backward calculations through the LLM during training and inference, because  CILMP operates solely on the extracted fixed representations from the LLM.
Therefore, these comparisons highlight the efficiency of our approach.
}

\subsection{{{Qualitative} Analysis}}
\begin{figure}[t!]
\centering
\includegraphics[width=0.47\textwidth]{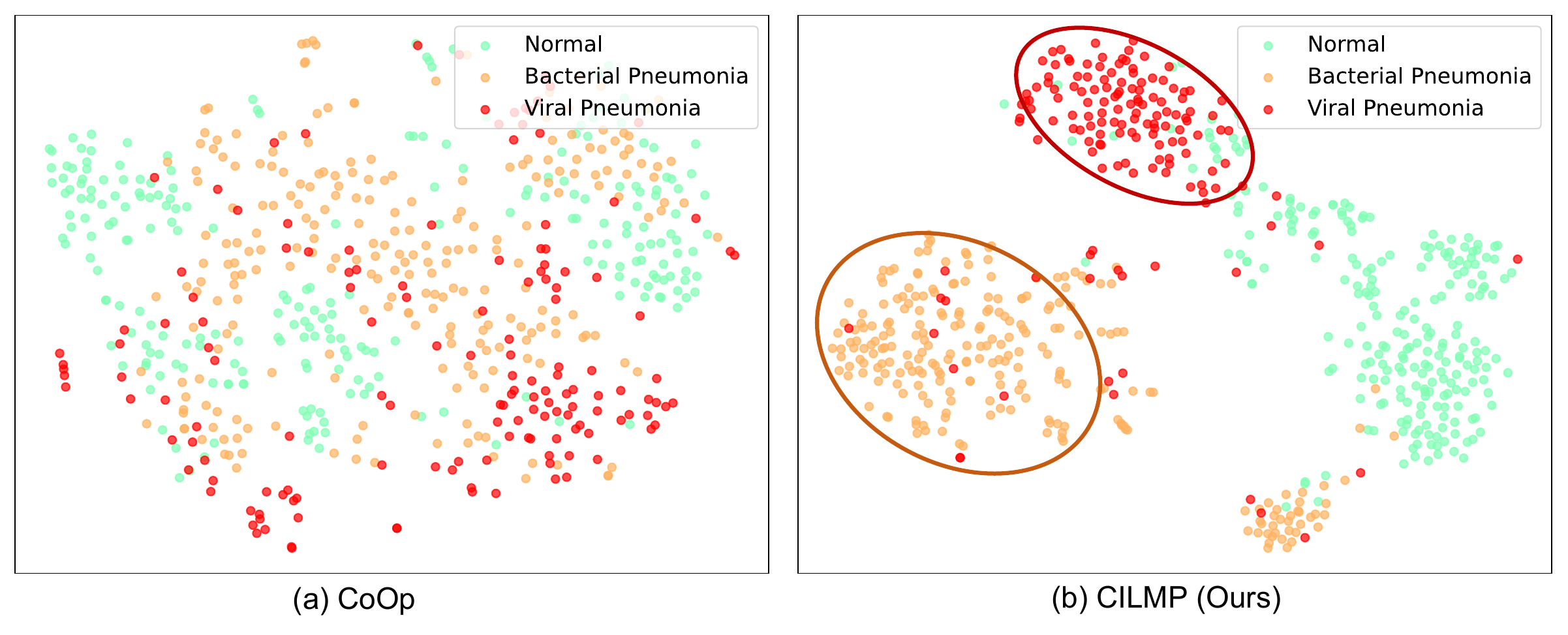}
\vspace{-0.5em}
\caption{{{Qualitative} analysis based on t-SNE}~\cite{van2008visualizing} {visualization. Compared to conventional prompt tuning, CILMP generates features that are more discriminative across classes.}
}
\vspace{-2em}
\label{fig:tsne}
\end{figure}

{
Fig.} \ref{fig:tsne} {presents a qualitative analysis of the feature space using t-SNE visualization}~\cite{van2008visualizing}. 
{We specifically examine the subtle differences in visual features between two challenging disease types: bacterial pneumonia and viral pneumonia.
The results indicate that traditional prompt tuning leads to ambiguous feature representation for these categories. 
This ambiguity arises because traditional methods rely solely on textual class names for differentiation, while bacterial and viral pneumonia share significant textual similarities. 
In contrast, our CILMP method demonstrates enhanced class-wise discrimination between the two categories, due to the incorporation of disease-adaptive prompts. 
This underscores the superiority of our approach.
However, the figure also reveals the presence of some outliers. For instance, several examples from the viral pneumonia class are found within the cluster of bacterial pneumonia. 
This shows a potential limitation of our method.
Modeling at the image feature level could be a viable solution to address the failure cases, and this lies beyond the scope of our current work and will be explored in future research.}

\begin{table}[t!]
\caption{Ablation study on the length of prefix/suffix of the intervention process. We simply set the prefix length equal to the suffix length. 
}
\label{tab:prefix}
\centering
\renewcommand\arraystretch{0.9}
\resizebox{0.48\textwidth}{!}{
\begin{tabular}{l|c|c|c|c} \toprule
\textbf{Length}      & \textbf{2}             & \textbf{4}             & \textbf{8}             & \textbf{16}            \\ \midrule
{ACC} [\%]   & 86.33$\pm$0.38 & {86.90$\pm$0.42} & 85.97$\pm$0.52 & 86.10$\pm$0.14  \\
{F1} [\%]    & 77.07$\pm$0.05 & 77.40$\pm$0.99 & 77.53$\pm$0.21 & 75.70$\pm$0.28   \\
{AUC} [\%]   & 97.73$\pm$0.09 & 97.90$\pm$0.00 & 97.73$\pm$0.12 & 97.63$\pm$0.05  \\
{Kappa} [\%] & 76.83$\pm$0.80 & 77.47$\pm$0.75 & 76.20$\pm$0.78 & 76.27$\pm$0.05  \\ \bottomrule        
\end{tabular}}
\end{table}

\begin{table}[t!]
\centering
\caption{Ablation study on the dimensionality of the linear subspace of the intervention function. 
}
\label{tab:linear-subspace}
\renewcommand\arraystretch{0.9}
\resizebox{0.49\textwidth}{!}{
\begin{tabular}{l|c|c|c|c} \toprule
\textbf{Dimension}      & \textbf{1}             & \textbf{4}             & \textbf{8}             & \textbf{16}            \\ \midrule
{ACC} [\%]   & 86.53$\pm$0.62 & 86.40$\pm$0.65 & {86.90$\pm$0.42} & 86.30$\pm$0.92 \\
{F1} [\%]    & 77.03$\pm$0.50 & 76.00$\pm$1.94 & 77.40$\pm$0.99 & {77.60$\pm$0.75} \\
{AUC} [\%]   & 97.90$\pm$0.08 & 97.83$\pm$0.05 & 97.90$\pm$0.00 & 97.80$\pm$0.14 \\
{Kappa} [\%] & 76.87$\pm$0.98 & 76.70$\pm$1.10 & 77.47$\pm$0.75 & 77.00$\pm$1.22 \\ \bottomrule        
\end{tabular}}
\end{table}

\begin{table}[t!]
\centering
\caption{{Ablation study on the prompt length.}
}
\label{tab:prompt-length}
\renewcommand\arraystretch{0.9}
{
\resizebox{0.49\textwidth}{!}{
\begin{tabular}{l|c|c|c|c} \toprule
\textbf{Length}      & \textbf{1}             & \textbf{4}             & \textbf{8}             & \textbf{16}            \\ \midrule
{ACC} [\%]  & 86.07$\pm$0.48 & 86.90$\pm$0.42 & {86.40$\pm$0.14} & 86.40$\pm$0.28  \\
{F1} [\%]   & 75.83$\pm$1.23 & 77.40$\pm$0.99 & 77.17$\pm$0.33  & 76.63$\pm$0.19  \\
{AUC} [\%]  & 97.63$\pm$0.21 & 97.90$\pm$0.00 & 97.83$\pm$0.05  & 97.93$\pm$0.24 \\
{Kappa} [\%] & 76.03$\pm$0.96 & 77.47$\pm$0.75 & 76.67$\pm$0.33  & 76.70$\pm$0.42 \\ \bottomrule        
\end{tabular}}
}
\vspace{-0.5em}
\end{table}

\section{Discussions} \label{sec:discuss}
\subsection{Employing more Large Language Models}
As large language models are crucial for integrating disease-specific knowledge into VLMs, {we assess the performance of more open-source LLMs, including LLaMA-2-7B}~\cite{touvron2023llama2}, {LLaMA-3-8B}~\cite{llama3modelcard}, {OpenBioLLM-8B}~\cite{OpenBioLLMs} {and MedLLaMA3-V20-8B}~\cite{medllama3}, with results presented in TABLE~\ref{tab:LLM}. 
{OpenBioLLM-8B and MedLLaMA3-V20-8B are esteemed as among the most powerful open-source medical LLMs, according to the Open Medical LLM Leaderboard}~\cite{medleaderboard}.

{The results indicate that LLaMA3-8B consistently outperforms LLaMA2-7B across all metrics, with improvements of approximately 0.4\% in accuracy, 0.9\% in F1 score, 0.1\% in AUC score, and 0.8\% in Kappa score.} 
This suggests that more advanced LLMs enhance performance due to their sophisticated architecture and comprehensive pre-training.

{Then, we {evaluate} how an LLM pre-trained specifically on medical data could enhance the performance of our framework. }
{The results in TABLE} \ref{tab:LLM} {{show} that: \textbf{(i)} Medical domain LLMs generally enhance performance on both the ISIC and DermaMNIST datasets. This improvement stems from their specialized training on medical data, fostering a deeper comprehension of diseases and offering more pertinent prior knowledge in contrast to general LLMs, such as LLaMA3.
Consequently, they generate more distinctive disease representations that aid in refining the visual language model prompt tuning.
\textbf{(ii)} Integration of these medical domain LLMs elevates CILMP's performance on the CPN-X-ray dataset, where CILMP initially lags behind other methods by a considerable margin, boosting accuracy from 96.20\% to 97.47\% and F1 score from 96.30\% to 97.53\%. Furthermore, on the challenging ODIR dataset, incorporating MedLLaMA3-V20-8B leads to a notable 2.17\% enhancement in F1 score.
These findings underscore the capability of medical domain LLMs to
enhance the quality of disease-specific representations, thereby augmenting the discriminability of prompts and diagnostic performance.

{Finally}, these results demonstrate the flexibility of our CILMP framework, which allows seamless integration with various LLMs without requiring modifications to the architectures. }
This adaptability positions our approach to accommodate future advancements in LLMs, thus ensuring continuous improvements in applicability.

\begin{table}[t]
\caption{{Experimental comparison based on various Large Language Models (general \& medical).}}
\centering
\renewcommand\arraystretch{0.95}
\resizebox{0.49\textwidth}{!}{
\begin{tabular}{l|cccc} \toprule
                   \textbf{LLM} & \textbf{ACC} [\%]  & \textbf{F1} [\%]   & \textbf{AUC} [\%] & \textbf{Kappa} [\%] \\ \midrule
\multicolumn{5}{l}{\textbf{\textit{ISIC:}}} \\
LLaMA2-7B \cite{touvron2023llama2}    & 86.47$\pm$0.54 & 76.53$\pm$1.13 & 97.83$\pm$0.09  & 76.70$\pm$0.96  \\
                             LLaMA3-8B \cite{llama3modelcard}    & 86.90$\pm$0.42 & 77.40$\pm$0.99 & 97.90$\pm$0.00  & 77.47$\pm$0.75    \\ 
OpenBioLLM-8B~\cite{OpenBioLLMs}  &
87.37$\pm$0.48 & 77.50$\pm$1.45 & 97.93$\pm$0.09 & 78.37$\pm$0.93
\\ 
MedLLaMA3-V20-8B~\cite{medllama3} & 87.33$\pm$0.09 & 78.63$\pm$0.24 & 97.80$\pm$0.14 & 78.33$\pm$0.25 \\    \midrule
 \multicolumn{5}{l}{\textbf{\textit{DermaMNIST:}}} \\
LLaMA2-7B \cite{touvron2023llama2}    &  87.00$\pm$0.24 & 77.80$\pm$1.42 & 98.13$\pm$0.12 & 75.13$\pm$0.24   \\
LLaMA3-8B \cite{llama3modelcard}  & 87.43$\pm$0.33 & 77.73$\pm$0.97 & 98.03$\pm$0.05 & 76.03$\pm$0.45      \\ 
OpenBioLLM-8B~\cite{OpenBioLLMs}  & 88.13$\pm$0.71 & 81.47$\pm$1.24 & 98.40$\pm$0.16 & 77.77$\pm$1.13 \\ 
MedLLaMA3-V20-8B~\cite{medllama3}  & 88.63$\pm$0.34 & 81.67$\pm$1.11 & 98.37$\pm$0.21 & 78.30$\pm$0.86 \\  \midrule
 \multicolumn{5}{l}{\textbf{\textit{CPN-X-ray:}}} \\
 LLaMA3-8B \cite{llama3modelcard}  & 96.20$\pm$0.33 & 96.30$\pm$0.33 &99.57$\pm$0.05 & 94.30$\pm$0.49  \\ 

 MedLLaMA3-V20-8B~\cite{medllama3}  & 97.47$\pm$0.12 & 97.53$\pm$0.17 & 99.77$\pm$0.05 & 96.20$\pm$0.22 \\  \midrule

\multicolumn{5}{l}{\textbf{\textit{ODIR:}}} \\
 LLaMA3-8B \cite{llama3modelcard}  & 68.47$\pm$0.87 & 57.93$\pm$1.32 &90.67$\pm$0.31 & 53.43$\pm$1.41    \\ 
 MedLLaMA3-V20-8B~\cite{medllama3}  & 69.03$\pm$0.75 & 60.10$\pm$1.45 & 91.43$\pm$0.31 & 54.73$\pm$1.11 \\
                            \bottomrule 
\end{tabular}}
\vspace{-0.5em}
\label{tab:LLM}
\end{table}




\subsection{Sensitivity to data-efficient scenarios}
\begin{figure}[t]
\centering
\includegraphics[width=0.37\textwidth]{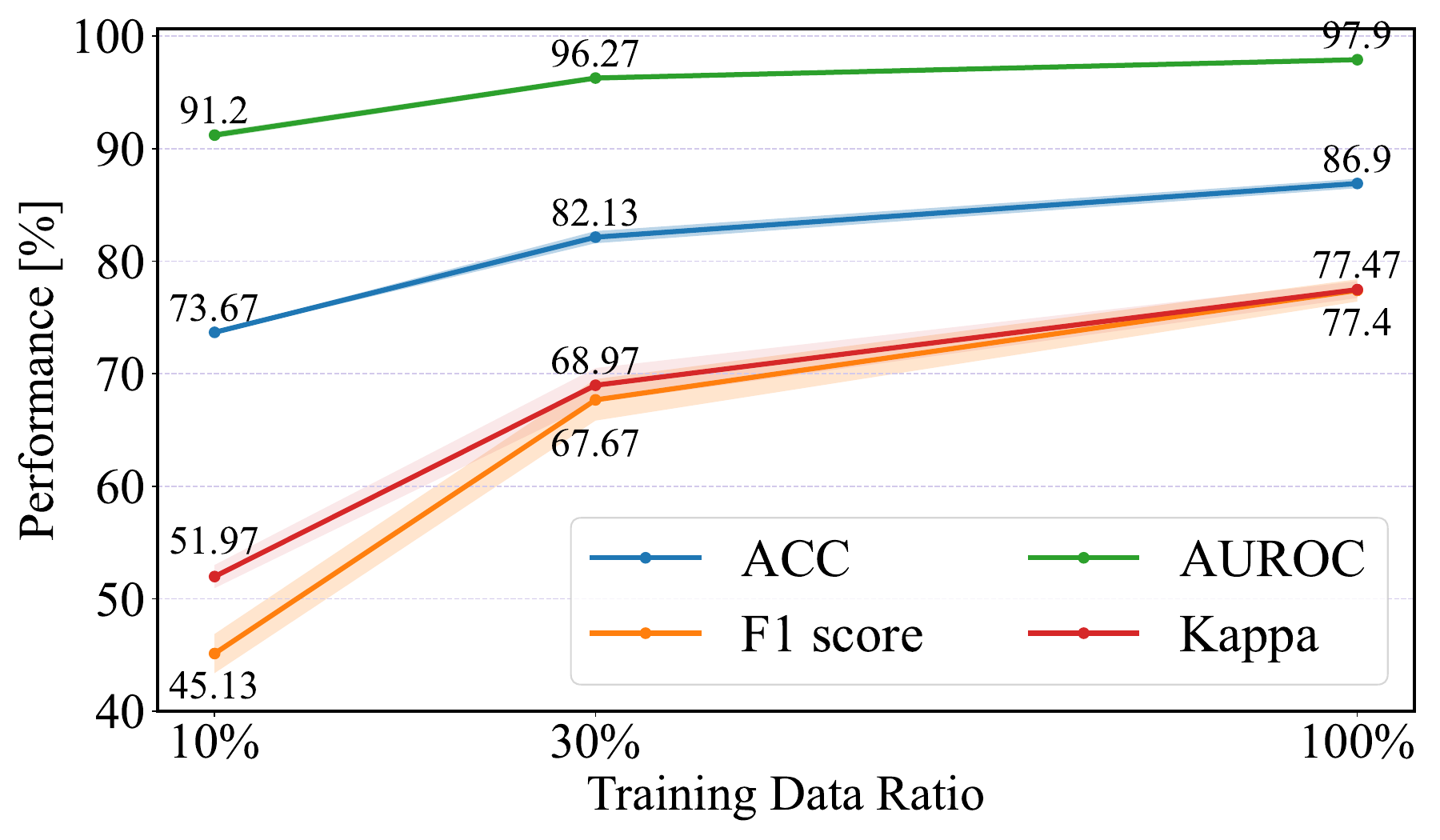}
\caption{Discussion on the sensitivity to data-efficient scenario.}
\label{fig:ratio}
\vspace{-1.4em}
\end{figure}

In many real-world applications, computing resources and the availability of data or labels are often limited. Thus, we examine the impact of data-efficient training scenarios on the performance of our CILMP method by testing its sensitivity to different training data ratios.
We conduct experiments using 10\%, 30\%, and 100\% of the available labeled training data with results presented in Fig.~\ref{fig:ratio}. 
At a 10\% label ratio, the framework achieves an accuracy of 73.67\%, an F1 score of 45.13\%, an AUC of 91.20\%, and a Kappa score of 51.97\%, indicating reasonable performance even with limited labeled data.
Increasing the label ratio to 30\% significantly enhances performance across all metrics. 
At the maximum label ratio of 100\%, the framework achieves its highest performance.
These results demonstrate that while the framework is robust in data-limited scenarios, it scales effectively with additional labeled data, achieving optimal performance when fully trained. This adaptability is crucial for practical applications where the availability of labeled data may vary, underscoring the framework’s potential for deployment in real-world settings.

\subsection{Sensitivity to class imbalance}


\begin{figure}[t]
\centering
\includegraphics[width=0.4\textwidth]{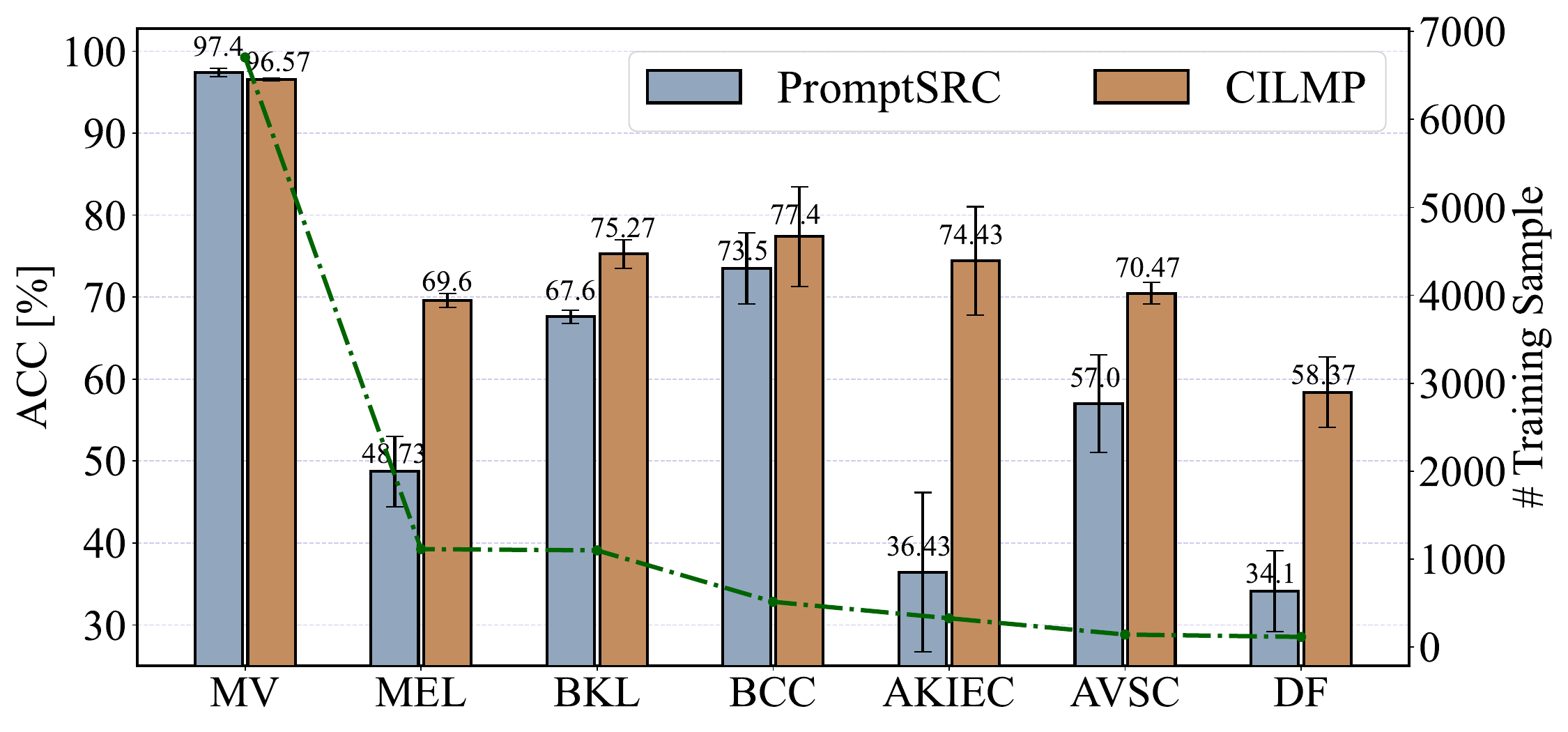}
\vspace{-0.5em}
\caption{Discussion on the sensitivity to the class imbalance issue. The accuracy [\%] over three random runs is reported.}
\vspace{-1.5em}
\label{fig:classimblance}
\end{figure}

Class imbalance is a prevalent issue in many real-world datasets, leading to biased models that perform poorly on minority classes.
To evaluate the robustness of our CILMP framework to label imbalance, we assess its performance on the ISIC 2018 dataset, which has varying numbers of training samples per class. 
The results, presented in Fig.~\ref{fig:classimblance}, compare our CILMP method against the PromptSRC method across seven classes: melanoma (MV), benign keratosis (MEL), benign keratosis (BKL), basal cell carcinoma (BCC), actinic keratosis (AKIEC), vascular lesion (AVSC), and dermatofibroma (DF).
Despite the imbalance observed, CILMP consistently shows robust performance, particularly excelling in minority classes. 
For instance, CILMP achieves an accuracy of 77.40\% for BCC, 74.43\% for AKIEC, 70.47\% for AVSC, and 58.37\% for DF, outperforming PromptSRC in these categories.
This robustness is attributed to CILMP's adaptive learning mechanisms, which generate instance-adaptive prompts that enhance learning on tail classes. In summary, the CILMP framework effectively handles class imbalance and limited labeled data, making it a versatile and reliable tool for real-world applications with uneven data distributions.

\subsection{{Limitations and Future Studies}}
{Despite the potential advantages offered by the proposed CILMP methodology, it still has some limitations in both technical and clinical aspects.}

\subsubsection{{Technical Limitation and Future Study}}
From a technical perspective, while CILMP does exhibit a reduction in training parameters compared to the fully fine-tuning paradigm, its implementation necessitates a non-trivial allocation of computational resources to facilitate the inference of a large language model for knowledge extraction. 
This requirement may present formidable obstacles in contexts characterized by severe computational constraints. 
Additionally, the current research has primarily focused on the application of CILMP in 2D medical image classification, leaving its potential in other domains—such as 3D imaging or video-based diagnostics—largely unexplored. 
Investigating the framework's efficacy in these areas could unlock its broader applicability.

Moreover, integrating advanced multi-modal fusion techniques, such as combining imaging data with patient and clinical information, presents an opportunity to improve the framework’s adaptability and diagnostic performance in real-world clinical settings. 
For instance, leveraging patient histories, laboratory results, or genomic data alongside imaging could enable a more comprehensive diagnostic approach.
Future studies should address these limitations to refine the CILMP framework, explore its applicability across diverse medical domains and modalities, and assess its performance in varied clinical settings to ensure its broader impact and generalizability.

\subsubsection{{Clinical Limitation and Future Study}}
{The proposed CILMP framework also faces challenges from a clinical perspective.
One of the key challenges lies in its interpretability and transparency in terms of clinical usage.
Specifically, CILMP involves the transfer of medical knowledge at the latent representation level, which remains opaque and less controllable. This lack of transparency can hinder clinical acceptance, since healthcare practitioners prioritize accountability, traceability, and explainability.
To address this limitation, future research should focus on developing methods to enhance the interpretability of latent knowledge transfers, incorporating explainable AI (XAI) techniques}~\cite{tjoa2020survey} {to provide clearer insights into the CILMP's decision-making process.}

{Another clinical limitation involves the challenge of auditing the clinical knowledge encoded within the latent space. 
Although experiments have shown that CILMP exhibits strong discriminative capabilities in distinguishing certain fine-grained diseases (\textit{e.g.}, viral and bacterial pneumonia) with feature-level evidence} {(Fig.}~\ref{fig:tsne}{),} {it remains challenging to audit whether the clinical knowledge encapsulated in the features aligns accurately with expert clinical understanding. 
This issue is particularly concerning in cases involving rare diseases or atypical presentations.
Future efforts could focus on establishing systematic auditing procedures}~\cite{degrave2025auditing} {to assess the alignment of latent representations with expert clinical insights, particularly for rare or complex conditions. }

{Lastly, clinical knowledge is often population-specific, with substantial data drift across datasets acquired from different regions, countries, or ethnic groups}~\cite{sahiner2023data}. 
{The knowledge transfer from pretrained LLMs poses a risk of perpetuating biased information and could hinder the practical application of the CILMP framework, as recent studies}~\cite{suenghataiphorn2025bias, omar2024socio} {have revealed the presence of pervasive bias in LLMs for various clinical applications.
To address this limitation, future investigations should focus on exploring methods to mitigate bias and promote fairness in healthcare applications. 
This could involve developing domain generalization techniques}~\cite{yoon2024domain} {that account for demographic variability to enhance the generalizability across diverse population groups. 
By addressing these clinical limitations, refinements can be made to the CILMP framework to ensure safer and more equitable utilization in clinical settings.}

\section{Conclusion}\label{sec:conclusion}


In this paper, we propose an LLM-based prompt tuning method for medical image analysis, leveraging medical knowledge in large language models to create disease-specific prompts for vision-language foundation models. An intervention function introduced bridges the LLM and VLM for knowledge transfer, while a conditional mechanism integrates the matching prior to generate instance-adaptive prompts. Extensive experiments across 11 diverse medical datasets demonstrate that our CILMP method outperforms recent state-of-the-art prompt tuning methods, showcasing its effectiveness.

\bibliographystyle{IEEEtran}
\bibliography{main}

\end{document}